\PassOptionsToPackage{unicode}{hyperref}
\PassOptionsToPackage{hyphens}{url}
\documentclass[
  11pt,
]{article}

\usepackage[margin=0.85in]{geometry}
\usepackage{amsmath,amssymb,amsfonts}
\usepackage{graphicx}
\usepackage{booktabs}
\usepackage{tabularx}
\usepackage{array}
\newcolumntype{Y}{>{\raggedright\arraybackslash}X}
\usepackage{float}
\usepackage{placeins}
\usepackage[font=small,labelfont=bf]{caption}
\usepackage{subcaption}
\usepackage[colorlinks=true,linkcolor=blue,citecolor=blue,urlcolor=blue]{hyperref}

\usepackage{xcolor}
\usepackage{iftex}
\ifPDFTeX
  \usepackage[T1]{fontenc}
  \usepackage[utf8]{inputenc}
  \usepackage{textcomp} % provide euro and other symbols
\else % if luatex or xetex
  \usepackage{unicode-math} % this also loads fontspec
  \defaultfontfeatures{Scale=MatchLowercase}
  \defaultfontfeatures[\rmfamily]{Ligatures=TeX,Scale=1}
\fi
\usepackage{lmodern}
\ifPDFTeX\else
\fi
\IfFileExists{upquote.sty}{\usepackage{upquote}}{}
\IfFileExists{microtype.sty}{% use microtype if available
  \usepackage[]{microtype}
  \UseMicrotypeSet[protrusion]{basicmath} % disable protrusion for tt fonts
}{}
\makeatletter
\@ifundefined{KOMAClassName}{% if non-KOMA class
  \IfFileExists{parskip.sty}{%
    \usepackage{parskip}
  }{% else
    \setlength{\parindent}{0pt}
    \setlength{\parskip}{6pt plus 2pt minus 1pt}}
}{% if KOMA class
  \KOMAoptions{parskip=half}}
\makeatother
\usepackage{graphicx}
\makeatletter
\newsavebox\pandoc@box
\newcommand*\pandocbounded[1]{% scales image to fit in text height/width
  \sbox\pandoc@box{#1}%
  \Gscale@div\@tempa{\textheight}{\dimexpr\ht\pandoc@box+\dp\pandoc@box\relax}%
  \Gscale@div\@tempb{\linewidth}{\wd\pandoc@box}%
  \ifdim\@tempb\p@<\@tempa\p@\let\@tempa\@tempb\fi% select the smaller of both
  \ifdim\@tempa\p@<\p@\scalebox{\@tempa}{\usebox\pandoc@box}%
  \else\usebox{\pandoc@box}%
  \fi%
}
\def\fps@figure{htbp}
\makeatother
\providecommand{\tightlist}{%
  \setlength{\itemsep}{0pt}\setlength{\parskip}{0pt}}
\usepackage{bookmark}
\IfFileExists{xurl.sty}{\usepackage{xurl}}{} % add URL line breaks if available
\hypersetup{
  hidelinks,
  pdfcreator={LaTeX via pandoc}}

\usepackage{etoolbox}
\patchcmd{\thebibliography}{\section*}{\subsection*}{}{}

\author{}
\date{}

\begin{document}

\section{Directing Open-Ended Evolution in Artificial Life via
Multi-Scale Path Divergence}\label{directing-open-ended-evolution-in-artificial-life-via-multi-scale-path-divergence}

\textbf{Mikhail Akhtyrchenko}\(^{1}\), \textbf{Mikhail I.
Katsnelson}\(^{2,4}\), \textbf{Andrey Ustyuzhanin}\(^{1,2,3}\)

\(^1\)Constructor University, Bremen, Germany. \(^2\)Constructor
Knowledge Labs, Bremen, Germany. \(^3\)Institute for Functional
Intelligent Materials, NUS, Singapore. \(^4\)Institute for Molecules and
Materials, Radboud University, Nijmegen, Netherlands.

\begin{center}\rule{0.5\linewidth}{0.5pt}\end{center}

% supports: [C1] [C2] [C5] -- abstract touches every claim retained in this version
\begin{abstract}
\noindent
Open-ended evolution (OEE) in artificial life is typically driven by
uninterpretable, black-box neural-network complexity metrics, leaving
life-like systems disconnected from physical theories of complexity.
We introduce MSPD (Multi-Scale Path Divergence, denoted \(D_P\)), a
renormalization-group-inspired scalar that quantifies the temporal
multiscale organization of heterogeneity in local transition laws.
MSPD is defined at the population level as a functional of the
realised trajectory and is computed as a windowed finite-resolution
estimator, with consistency between the two stated as a proposition.
The metric is an explicit formula and plays a dual role: as a
gradient-free fitness function and as a post-hoc analytical lens on
any simulation that exposes local transition laws.
On a Flow-Lenia substrate we establish three claims.
(C1) Under fixed-context replay of the exact pathwise objective,
MSPD-optimized parameters score higher than matched random
parameters from the same substrate.
(C2) Along optimized trajectories, states whose local transition
laws are more heterogeneous yield larger future divergence than
matched, less-heterogeneous states under exact-state stochastic
continuations, so the metric tracks the intrinsic dynamics of those
trajectories rather than injected noise.
(C5) Higher MSPD corresponds to stronger scale-dependent frustration:
high-complexity systems exhibit larger differences between the
dynamics expressed at different spatial extents, linking MSPD
directly to the frustration criterion of biological complexity in
the sense of Vanchurin et al.~\cite{vanchurin2022toward}.
All three transfer beyond the primary Flow-Lenia substrate to
Life-like cellular automata and Particle Life++, indicating that MSPD
is not specific to a single substrate.
A single explicit formula thus both \emph{directs} open-ended
evolution and provides a principled bridge to the physics of
complexity that black-box drivers do not.
\end{abstract}

\begin{center}\rule{0.5\linewidth}{0.5pt}\end{center}

% supports: [C5] + cross-substrate transfer (this section also lists the contribution bullets)
\subsection{1 Introduction}\label{introduction}

The origin of complexity, and first of all biological complexity,
is one of the main challenges of contemporary science, and
many efforts have been applied to deal with it. Trying to make
a more or less complete list of various approaches is a hopeless
task, but we should at least mention numerous attempts to
establish relations between concepts and methods of statistical
physics and the problem of life and its evolution~\cite{laughlin1,laughlin2,gavrilets2004,smith2008thermo,goldenfeld2011life,katsnelson2018towards,wolf2018physical,vanchurin2022toward,vanchurin2022thermo}. It is clear that such a
complicated problem should be attacked from different sides.
One direction is more traditional: the analysis of true biological information
in terms of statistical physics, including glassiness~\cite{laughlin1,laughlin2}, frustration in general~\cite{wolf2018physical,katsnelson2018towards}, percolation~\cite{gavrilets2004,katsnelson2018towards},
emergent phenomena and universality classes~\cite{goldenfeld2011life}, and, last but not
least, analogies with machine learning~\cite{vanchurin2022toward,vanchurin2022thermo}. Another direction is
Artificial Life (ALife), that is, building in silico models with relatively simple
rules but complicated behaviour, where everything is in principle under control, which
gives us hope to better understand the road from simplicity to complexity. The approach
is intended to be complementary to the biological one but, hopefully, in the end should
tell us something important about real living systems and their evolution. The question,
however, is how to formally define complexity.

Again, it looks like a hopeless task to mention all existing approaches to
the formal definition of complexity. Some of them are related to the
idea that structural complexity is determined by scaling properties of the
system and its hierarchical organization, that is, by the dissimilarity between
the images of the system considered at different scales. This idea has appeared
many times in various contexts~\cite{wolpert2007,lindeberg2008,broido2019}.
Probably its most suitable formalization was suggested in~\cite{bagrov2020multiscale}
as ``multiscale structural complexity'' (MSSC). The definition is computationally
simple and, at the same time, has been demonstrated to be useful in many different
fields, including analysis of the complexity of quantum states and detection of
quantum phase transitions~\cite{sotnikov2022} and analysis of human visual
perception~\cite{kravchenko2026}. However, this definition is static and,
to be applied to an evolving system, needs a modification adding not only
spatial but also temporal scales.

A central goal of ALife is open-ended evolution---the sustained
generation of novel forms and behaviors~\cite{stanley2019why,lehman2011abandoning,soros2014identifying,pugh2016quality}. Neural-network-based
open-endedness metrics drive this effectively but are black boxes: they
yield results without principles and cannot be connected to physical
theories of complexity. MSSC offered a physics-grounded alternative
for static spatial patterns, but life is inherently dynamic. We extend
MSSC temporally, replacing the spatial coarse-graining scale \(\lambda\)
with an observational window \(W\), and ask: can a single explicit scalar
simultaneously optimize for complexity \emph{and} reveal its physical basis?
Thus, here we replace MSSC with Multi-Scale Path Divergence (MSPD), which
can in a sense be considered as its dynamical analogue and generalization.
Its explicit definition is given below.

% supports: cross-substrate transfer -- introduces the cross-substrate experimental scope
We validate MSPD in Flow-Lenia~\cite{plantec2023flow}, a mass-conservative
continuous cellular automaton in the Lenia family~\cite{chan2019lenia}
supporting diverse life-like gliders, and transfer experiments
to two qualitatively different substrates: Life-like cellular automata
(discrete-state lattices with totalistic update rules) and Particle
Life++ (neural-network-mediated pairwise particle interactions). The
discrete CA substrate exercises the categorical transition-law form of
MSPD, while Particle Life++ shares with Flow-Lenia the Lagrangian
displacement form and connects to a broader Lagrangian thread in ALife
exemplified by Particle Lenia~\cite{mordvintsev2022particle}.

Here we apply this new metric to the analysis of some general properties
of life formulated within the analogy between biological evolution and
multiscale learning~\cite{vanchurin2022toward,vanchurin2022thermo}, and especially to the
role of frustration, which has been suggested as the main cause of complexity
not only in physical but also in biological systems~\cite{wolf2018physical}.
% supports: [C5] -- scale-dependent frustration via constrained-rollout intervention
A key empirical result is that higher MSPD corresponds to stronger
\emph{scale-dependent frustration}: in optimized systems, the dynamics
expressed under an early spatial constraint differ from the unconstrained
dynamics by more than would be expected from ordinary seed-to-seed
variation. This intervention-based measurement links MSPD directly to
the frustration criterion of biological complexity in the sense of
Vanchurin et al.~\cite{vanchurin2022toward} (building on the spin-glass
framing of biological complexity by Wolf et al.~\cite{wolf2018physical}),
without the metric being optimised for frustration.

\begin{figure}[!ht]
\centering
\pandocbounded{\includegraphics[keepaspectratio,alt={Pipeline schematic}]{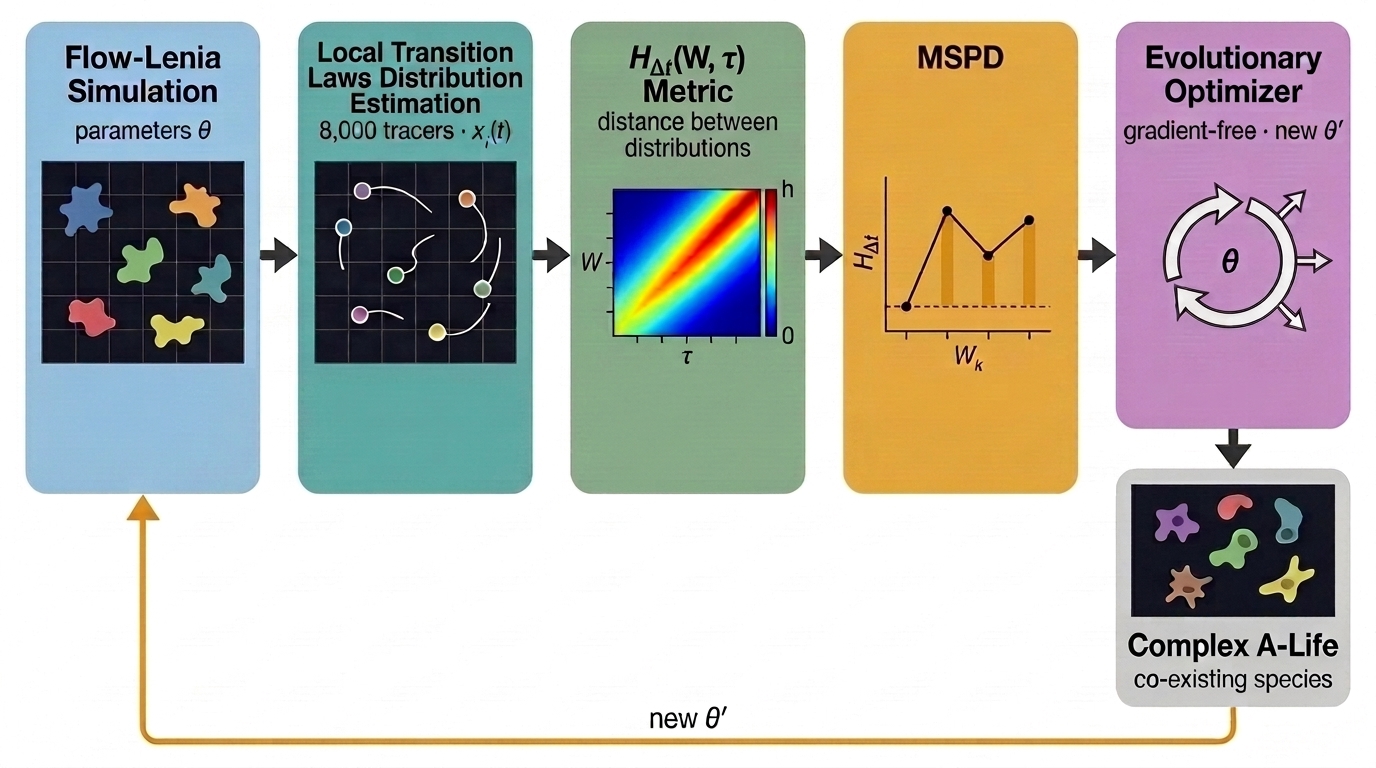}}
\caption{Pipeline: Lagrangian tracking \(\to\) \(H_{\Delta t}\)
heatmap \(\to\) MSPD aggregation \(\to\) evolutionary loop \(\to\)
complex simulation output.}
\label{fig:pipeline}
\end{figure}

Figure~\ref{fig:pipeline} summarises the full pipeline: Lagrangian
particle tracking, the per-window pairwise dissimilarity \(\Delta H\),
the cross-window aggregation that yields \(D_P\), and the evolutionary
loop that closes back onto the simulation.

\textbf{Contributions.}
\begin{itemize}
\tightlist
\item
  We introduce MSPD, a temporal extension of spatial MSSC defined as
  a functional of the realised trajectory via local transition laws,
  with a consistent windowed finite-resolution estimator. % framing
\item
  Used as a fitness function and as a lens, MSPD separates optimized
  parameters from matched random controls on the same substrate, and
  high-$\Delta H$ states are linked to higher instability under
  external interventions. % [C1] [C2]
\item
  MSPD-optimized systems exhibit elevated \emph{scale-dependent
  frustration} under matched constrained-rollout interventions,
  linking MSPD to the Vanchurin et al. frustration criterion of
  biological complexity without being optimised for it. % [C5]
\item
  The protocol transfers beyond the primary Flow-Lenia substrate:
  C1, C2 and C5 also hold on Life-like cellular automata and
  Particle Life++. % cross-substrate transfer
\end{itemize}

\begin{center}\rule{0.5\linewidth}{0.5pt}\end{center}

\subsection{2 Background and Related Work}\label{background-and-related-work}

% framing for: paper-wide (RG-grounded interpretable complexity)
\textbf{2.1 MSSC and renormalization-grounded complexity.} Spatial
Multi-Scale Structural Complexity~\cite{bagrov2020multiscale} defines
the complexity of a pattern as the cumulative cross-scale dissimilarity
under successive Kadanoff--Wilson/RG (renormalization group) \cite{wilson1974,mabook} coarse-graining: a system is structurally
complex to the extent that its representations at adjacent spatial
resolutions disagree. Unlike Kolmogorov-style~\cite{livitanyi2008} or
computational-mechanics measures~\cite{shalizi2001computational,crutchfield2012}, MSSC is computed directly from
continuous field data without abstract ensembles or
$\epsilon$-machine inference; it peaks for hierarchically nested
structures (spin spirals, biological tissues) and drops to near-zero
for both white noise and uniform fields. Subsequent
work~\cite{kravchenko2026} shows that MSSC also tracks human
perceptual judgements of visual complexity once the highest- and
lowest-frequency scales are filtered out --- an empirical observation
that motivates the log-spaced window choice we adopt for MSPD
in~\S3.2. MSPD inherits this RG architecture intact: the
coarse-graining scale $\lambda$ is replaced by an observation window
$W$, and the cross-scale dissimilarity is computed in
trajectory-distribution space rather than image space.

% framing for: cross-substrate transfer -- justifies why Flow-Lenia/Boids/Particle Life all admit MSPD
\textbf{2.2 Continuous ALife substrates.}
Lenia~\cite{chan2019lenia} generalised Conway's Game of Life to a
continuous space--time--state framework, replacing discrete rule
tables with integro-differential update rules and producing a
taxonomy of self-maintaining gliders. Flow-Lenia~\cite{plantec2023flow}
adds strict mass conservation: patterns must compete for finite
material resources, which is the thermodynamic precondition for
ecosystemic dynamics and the substrate on which we run our primary
experiments. Particle Lenia~\cite{mordvintsev2022particle}, Particle
Life, and the classical Boids model take a strictly Lagrangian view,
modelling matter as discrete interacting particles rather than as
scalar fields. MSPD applies uniformly across these substrates because
each one exposes explicit particle trajectories --- whether as native
state in particle systems or as passive tracers injected into a
Flow-Lenia mass field.

% framing: general -- positions MSPD against black-box NN-OEE drivers
% (head-to-head richness comparison deferred; see backlog C4)
\textbf{2.3 Open-endedness drivers.}
Open-ended evolution is a recognised grand challenge of
ALife~\cite{stanley2019why}: objective-driven evolutionary algorithms
collapse on deceptive landscapes, motivating a family of alternatives
that target diversity or novelty instead of a single objective.
Lehman \& Stanley~\cite{lehman2011abandoning} introduced Novelty
Search, abandoning fitness in favour of behavioural novelty; Soros \&
Stanley~\cite{soros2014identifying} enumerated necessary conditions for
OEE in cellular substrates; and Pugh, Soros \&
Stanley~\cite{pugh2016quality} formalised Quality-Diversity (QD),
which maintains a tessellated archive of high-performing,
behaviourally diverse solutions. Two recent ALife-specific drivers
are particularly relevant:
Leniabreeder~\cite{faldor2024leniabreeder} applies QD to Lenia using
either hand-crafted descriptors or autoencoder-latent embeddings, and
ASAL~\cite{kumar2024automating} uses a vision--language foundation
model as the interestingness judge over simulation videos. The shared
limitation of all of these drivers is that their notion of
``interesting'' is either hand-tuned or learned, and is therefore not
connected to physical theories of complexity --- the gap MSPD addresses
with an explicit, formula-based scalar.

% framing for: [C5] -- introduces the frustration criterion MSPD will recover
\textbf{2.4 Physical theories of life.}
Wolf, Katsnelson \& Koonin~\cite{wolf2018physical} following the previous
observations of Laughlin and others~\cite{laughlin1,laughlin2} argue that
biological systems are forms of \emph{frustrated matter}: analogues of
spin glasses, in which competing interactions prevent the system from
settling into any single global equilibrium and instead produce a
rugged landscape with a broad distribution of relaxation timescales.
Vanchurin et al.~\cite{vanchurin2022toward} extend this picture into a
multilevel-learning theory of evolution and identify frustration /
non-ergodicity as an operational signature of living systems. This is
the physical criterion that MSPD-optimised systems satisfy
\emph{post-hoc} (\S4, C5) without ever being optimised for it --- the
bridge that turns MSPD from a fitness function into a diagnostic of
life-likeness.

% framing for: methodological -- justifies the SW_1 distance choice
\textbf{2.5 Trajectory and Wasserstein analysis.}
Comparing distributions of particle trajectories requires a
topology-aware divergence: KL and MMD lose meaning or scale poorly
when distributions have disjoint or near-disjoint supports, which is
generic in trajectory data because each particle traces out a
different region of phase space. The Wasserstein-1 metric (Earth
Mover's Distance) supplies the right notion of distance --- the
minimal cost of transporting one distribution onto another, accounting
for the underlying metric of the space. Full $W_1$ scales cubically in
the number of samples; we therefore use its sliced approximation
$\mathrm{SW}_1$, which projects distributions onto random one-dimensional
directions and averages closed-form 1-D Wasserstein distances.
$\mathrm{SW}_1$ is metric-equivalent to $W_1$ on bounded supports
while costing $O(n \log n)$ per projection, which is what makes the
per-particle, per-window comparison over thousands of tracers
tractable.

\begin{center}\rule{0.5\linewidth}{0.5pt}\end{center}

\subsection{3 Method}
\label{method}

MSPD measures the temporal organization of heterogeneity in local dynamics.  The construction begins with a formal model for systems whose states are configurations of local degrees of freedom.  From the transition law of such a system we obtain local transition laws.  Their pairwise variation defines state-level dynamical heterogeneity.  Along a realised trajectory this gives a scalar heterogeneity trace, and MSPD measures how this trace changes across temporal coarse-graining scales.

The section is organized as follows.  We first define the configuration model and the local transition laws induced by the dynamics.  We then define state-level heterogeneity and the pathwise MSPD functional.  The final part derives the computable estimator as a windowed finite-resolution approximation to the theoretical quantity and states the consistency relation between the two.

\subsubsection*{3.1 Configuration model for local dynamics}
\label{method:system-model}

We model the observed system by a configuration space.  Let
\((\mathcal I,\mathcal B_{\mathcal I})\) be a measurable index space for local degrees of freedom.  An element \(i\in\mathcal I\) is one local component of the system: for example a lattice site, a particle, an agent, a spatial sample, or a sampled material element.  Let \((E,\mathcal B_E)\) be the local state space.  A configuration is a map
\[
  x:\mathcal I\to E,
\]
where \(x(i)\) is the local state of degree of freedom \(i\).  We assume that the admissible configurations form a measurable space \((\mathcal X,\mathcal B_{\mathcal X})\) with \(\mathcal X\subseteq E^{\mathcal I}\).  The sigma-algebra \(\mathcal B_{\mathcal X}\) is generated by the coordinate maps
\[
  \operatorname{ev}_i:\mathcal X\to E,
  \qquad
  \operatorname{ev}_i(x)=x(i).
\]

The time-\(\Delta t\) dynamics is a Markov kernel
\[
  P_{\Delta t}:\mathcal X\times\mathcal B_{\mathcal X}\to[0,1].
\]
For each current configuration \(x\), \(P_{\Delta t}(x,\cdot)\) is the probability law of the next configuration.  In deterministic dynamics, this kernel is induced by an update map \(\Phi_{\Delta t}:\mathcal X\to\mathcal X\), where
\[
  P_{\Delta t}(x,\cdot)=\delta_{\Phi_{\Delta t}(x)}.
\]
The notation \(P_{\Delta t}(x,dy)\) denotes integration with respect to the probability measure over future configurations \(y\).

The configuration model also requires a way to average over local degrees of freedom.  We write
\[
  m_x\in\mathcal P(\mathcal I)
\]
for the sampling measure in configuration \(x\).  This measure encodes which local components contribute to the measured dynamical heterogeneity and with what weight.  Uniform weighting over sites, uniform weighting over agents, activity weighting, and mass weighting define different dynamical observables.  Thus \(m_x\) is part of the formal measurement specification: it determines the contribution of local degrees of freedom to the system-level average.

The objects \(\mathcal I\), \(E\), \(\mathcal X\), \(P_{\Delta t}\), \(m_x\), and the transition-law distance introduced below specify the local dynamical observable to which MSPD is applied.  They are not numerical details of the estimator.  They define the formal description of the dynamics being measured.

\subsubsection*{3.2 Local transition laws}
\label{method:local-transition-laws}

The key local object is the transition law induced by the system dynamics at one local degree of freedom.  For a current configuration \(x\) and \(i\in\mathcal I\), define
\[
  G_{x,i}:\mathcal X\to E\times E,
  \qquad
  G_{x,i}(y)=(x(i),y(i)).
\]
This map returns the before-after local state pair at \(i\) when the next configuration is \(y\).  The local transition law is the pushforward measure
\begin{equation}
T_{\Delta t}^{i}(x)
  =
  (G_{x,i})_{\#}P_{\Delta t}(x,\cdot)
  \in\mathcal P(E\times E).
  \label{eq:local-transition-law}
\end{equation}
Equivalently, for measurable \(B\subseteq E\times E\),
\[
  T_{\Delta t}^{i}(x)(B)
  =
  P_{\Delta t}\left(x,
  \{y\in\mathcal X:(x(i),y(i))\in B\}
  \right).
\]
The law \(T_{\Delta t}^{i}(x)\) is obtained from the transition kernel and the coordinate \(i\).  No additional feature map is introduced.  In deterministic dynamics,
\[
  T_{\Delta t}^{i}(x)
  =
  \delta_{(x(i),\Phi_{\Delta t}(x)(i))}.
\]

When \(E\) is a vector space, transition-pair laws can be represented as laws of local increments.  Let
\[
  \nabla_{\Delta t}(e,e')=\frac{e'-e}{\Delta t}.
\]
The local increment law is
\begin{equation}
V_{\Delta t}^{i}(x)
  =
  (\nabla_{\Delta t})_{\#}T_{\Delta t}^{i}(x)
  \in\mathcal P(E).
  \label{eq:local-increment-law}
\end{equation}
For moving particles, agents, or transported material samples, this is a displacement or velocity law.  For discrete local states, such as spins or finite-state automata, the transition-pair law \(T_{\Delta t}^{i}(x)\) remains the appropriate object.

Although the examples considered later have local or short-range update rules, the definition of \(T_{\Delta t}^{i}(x)\) does not require strict locality of interactions.  The transition law may depend on the full configuration \(x\).  Long-range interactions are represented through their effect on the local transition laws.  The version used here compares one-component transition laws.  If the object of interest is the heterogeneity of joint transitions, the same construction can be applied to finite blocks \(B\subset\mathcal I\), replacing \(i\) by \(B\) and using
\[
  T_{\Delta t}^{B}(x)
  =
  \left(y\mapsto (x|_B,y|_B)\right)_{\#}P_{\Delta t}(x,\cdot).
\]

\subsubsection*{3.3 State-level dynamical heterogeneity}
\label{method:state-heterogeneity}

Let \(D\) be a distance or divergence between probability measures on \(E\times E\).  The state-level dynamical heterogeneity is
\begin{equation}
\mathcal H_{\Delta t}(x)
  =
  \iint_{\mathcal I\times\mathcal I}
  D\left(T_{\Delta t}^{i}(x),T_{\Delta t}^{j}(x)\right)
  \,m_x(di)m_x(dj).
  \label{eq:state-heterogeneity-transition}
\end{equation}
If increment laws are used, the corresponding form is
\begin{equation}
\mathcal H_{\Delta t}(x)
  =
  \iint_{\mathcal I\times\mathcal I}
  D\left(V_{\Delta t}^{i}(x),V_{\Delta t}^{j}(x)\right)
  \,m_x(di)m_x(dj),
  \label{eq:state-heterogeneity-increment}
\end{equation}
where \(D\) is now a distance between probability measures on \(E\).

The definition compares the transition laws of two local degrees of freedom sampled from the same state.  If all local degrees of freedom have the same transition law \(m_x\)-almost everywhere, then \(\mathcal H_{\Delta t}(x)=0\).  If different parts of the state have different local transition laws, then \(\mathcal H_{\Delta t}(x)>0\).

A simple mixture illustrates the value of \(\mathcal H_{\Delta t}\).  Suppose the local degrees of freedom split into two transition types with laws \(L_1\) and \(L_2\).  If the first type has mass \(p\) under \(m_x\) and the second has mass \(1-p\), then
\begin{equation}
\mathcal H_{\Delta t}(x)=2p(1-p)D(L_1,L_2).
  \label{eq:two-type-H}
\end{equation}
For a homogeneous state, \(p=0\) or \(p=1\), the heterogeneity is zero.  For a mixed state it is positive whenever the two transition laws differ.  For example, if local components are Brownian particles with two diffusion laws, then \(L_1\) and \(L_2\) are the corresponding Gaussian increment laws.  A single-diffusivity Brownian system has zero state heterogeneity, while a two-diffusivity mixture has positive heterogeneity proportional to \(2p(1-p)\) and to the distance between the two Gaussian transition laws.

\subsubsection*{3.4 Pathwise and system-level MSPD}
\label{method:pathwise-mspd}

State-level heterogeneity is an instantaneous quantity.  To obtain a dynamical complexity for a system, we study how this quantity evolves along trajectories.  A system is specified together with an evaluation protocol: an initial-condition law and any internal randomness of the update.  Together with the transition kernel, this induces a probability law \(\mathbb P_{\mathcal S}\) on realised trajectories \(\xi=(x_t)_{t\ge0}\) in \(\mathcal X\).  The coordinate process is \(X_t(\xi)=x_t\).

A realised trajectory \(\xi\) first defines the raw heterogeneity trace
\begin{equation}
g_\xi(t)=\mathcal H_{\Delta t}(X_t(\xi)).
  \label{eq:pathwise-raw-heterogeneity}
\end{equation}
The score is applied to a nonnegative measurement trace
\begin{equation}
h_\xi(t)=\phi(g_\xi(t)),
  \label{eq:pathwise-heterogeneity}
\end{equation}
where \(\phi\) is fixed as part of the measurement specification.  In the ideal case \(\phi\) may be the identity; in finite computations it is the same positive/floor preprocessing used on the empirical heterogeneity estimates.  The average level of \(h_\xi\) is the heterogeneity amplitude.  MSPD measures a different property: the temporal organization of this heterogeneity across scales.

Let \(\mathcal A_r\) be temporal coarse-graining at scale \(r\), and set
\[
  h_r=\mathcal A_rh.
\]
A trace is multiscale when its coarse-grained versions change nontrivially as the observation scale is varied.  Temporal scale is multiplicative: changing the scale from \(10\) to \(20\) and from \(100\) to \(200\) represents the same relative change of resolution.  We therefore use the logarithmic scale derivative
\[
  \partial_{\log r}h_r=r\partial_rh_r.
\]
For a probability measure \(q\) over temporal scales and a floor \(\eta>0\), the pathwise MSPD of a nonnegative trace \(h\) is
\begin{equation}
\operatorname{MSPD}[h_\xi]
  =
  \int
  \frac{
    \|\partial_{\log r}h_r\|_{2,t}^{2}
  }{
    \|h_r\|_{2,t}^{2}+\eta^2
  }
  \,q(d\log r).
  \label{eq:pathwise-mspd}
\end{equation}
The numerator measures the scale sensitivity of the heterogeneity trace.  The denominator normalizes by the signal power at that scale and prevents small residual traces from producing large relative scores.

Two limiting cases are immediate.  If \(h(t)\equiv0\), then \(h_r\equiv0\) for every \(r\), so \(\operatorname{MSPD}[h]=0\).  If \(h(t)\equiv c\) for a constant \(c>0\), then every coarse-grained trace is again constant and \(\partial_{\log r}h_r=0\).  Thus \(\operatorname{MSPD}[h]=0\) even though the heterogeneity amplitude is positive.  Stationary heterogeneity and temporally organized heterogeneity are therefore separated by construction.

The system-level quantity is obtained by applying the pathwise functional first and then averaging over realised trajectories:
\begin{equation}
\operatorname{MSPD}(\mathcal S)
  =
  \mathbb E_{\xi\sim\mathbb P_{\mathcal S}}
  \left[\operatorname{MSPD}[h_\xi]\right].
  \label{eq:system-level-mspd}
\end{equation}
This order matters because the pathwise functional is nonlinear.  Averaging heterogeneity traces before applying MSPD can cancel temporal structure across trajectories.  For ergodic systems, a long trajectory estimates a typical pathwise value.  For non-ergodic systems, the value is tied to the specified distribution of initial conditions and seeds.

\subsubsection*{3.5 Windowed finite-resolution target}
\label{method:windowed-target}

The state-level quantity \(\mathcal H_{\Delta t}(x)\) is defined from population transition laws at a configuration.  In simulations these transition laws are not observed directly.  A single deterministic transition provides one before-after sample for each local component, while \(\mathcal H_{\Delta t}\) compares transition laws.  The computable quantity is therefore introduced through an intermediate population object: a finite-resolution, windowed target.  This target is still defined at the population level, but it uses the same temporal resolution and observation windows as the estimator.

Let
\[
  I_k=[a_k,a_k+s]
\]
be an observation window of duration \(s\), and let
\[
  I_{k,\tau}=\{u\in I_k:u+\tau\in I_k\}
\]
be the admissible set of start times for a finite-difference transition at lag \(\tau\).  For a realised trajectory \(\xi\), let \(Z_i^\xi(t)\) denote the local state or local coordinate observed at degree of freedom \(i\) and time \(t\).  For vector-valued local observations, define the windowed finite-resolution transition law
\begin{equation}
L_{i,k,\tau}^{\xi}
  =
  \left(u\mapsto
    \frac{Z_i^\xi(u+\tau)-Z_i^\xi(u)}{\tau}
  \right)_{\#}\operatorname{Unif}(I_{k,\tau}).
  \label{eq:window-law}
\end{equation}
For systems whose local state is discrete or non-vectorial, the map inside the pushforward is replaced by the transition-pair map
\[
  u\mapsto (Z_i^\xi(u),Z_i^\xi(u+\tau)).
\]
Thus \(L_{i,k,\tau}^{\xi}\) is the population law of local transitions observed over the window \(I_k\) at temporal resolution \(\tau\).

\paragraph{Estimator resolution.}
The pair \((s,\tau)\) fixes the resolution at which local transition laws are measured from a finite rollout.  It is selected from the simulation and logging protocol, not from the downstream claim being tested.  Small lags can make the finite difference
\[
  \frac{Z_i^\xi(u+\tau)-Z_i^\xi(u)}{\tau}
\]
dominated by jitter, numerical noise, or short-scale stochastic motion.  Small windows give too few samples to estimate the law \(L_{i,k,\tau}^{\xi}\) reliably.  Conversely, very large lags or windows mix distinct local regimes and remove fast transitions.  We therefore use an admissible scale range rather than arbitrary values of \(s\) and \(\tau\).

A scale pair is admissible when the number of transition samples
\[
  m_k(s,\tau)=|I_{k,\tau}|
\]
exceeds a fixed minimum, neighboring lag estimates have reached a stable regime, and split-window estimates do not vary more than the between-carrier heterogeneity they are meant to resolve.  One practical diagnostic for lag stability is
\[
  R_\tau(s)=\operatorname{median}_{k}
  D\!\left(\bar L_{k,\tau}^{\xi},\bar L_{k,\tau_+}^{\xi}\right),
  \qquad
  \bar L_{k,\tau}^{\xi}=\int L_{i,k,\tau}^{\xi}\,m_k^\xi(di),
\]
where \(\tau_+\) is the next lag in the candidate grid.  The lower admissible lag is chosen after the sharp jitter-dominated changes in \(R_\tau(s)\) have subsided.  A practical diagnostic for window stability is a split-window discrepancy,
\[
  E_s(\tau)=\operatorname{median}_{i,k}
  D\!\left(L_{i,k,\tau}^{\xi,\mathrm{first}},L_{i,k,\tau}^{\xi,\mathrm{second}}\right),
\]
where the two laws are estimated from the first and second halves of the same window.  The window length is chosen large enough for this sampling error to be small, but not so large that the two halves systematically represent different dynamical regimes.  The same calibrated admissible-scale protocol is used for optimized systems and controls.

The corresponding windowed heterogeneity target is
\begin{equation}
H_{k,\tau}^{\xi}
  =
  \iint
  D\left(L_{i,k,\tau}^{\xi},L_{j,k,\tau}^{\xi}\right)
  \,m_k^\xi(di)m_k^\xi(dj),
  \label{eq:windowed-H}
\end{equation}
where \(m_k^\xi\) is the sampling law over local degrees of freedom on the same window.  Equation~\eqref{eq:windowed-H} is not an empirical estimator.  It is the population heterogeneity associated with the finite observation window and finite transition lag.

The link to the ideal trace is a local-stationarity approximation.  If local transition statistics do not vary substantially inside \(I_k\), then the raw windowed target approximates the local temporal average of the raw heterogeneity trace:
\begin{equation}
H_{k,\tau}^{\xi}
  \approx
  \frac{1}{|I_k|}
  \int_{I_k}g_\xi(u)\,du.
  \label{eq:window-approx-local-average}
\end{equation}
The trace fed into MSPD is the preprocessed trace \(h_\xi=\phi(g_\xi)\).  The corresponding population windowed trace is
\begin{equation}
  h_{k,\tau}^{\star}
  =
  \phi(H_{k,\tau}^{\xi}),
  \qquad k=1,\ldots,K.
  \label{eq:population-windowed-trace}
\end{equation}
Under the same local-stationarity assumption, \(h_{k,\tau}^{\star}\) approximates the local average of the preprocessed trace,
\begin{equation}
h_{k,\tau}^{\star}
  \approx
  \frac{1}{|I_k|}
  \int_{I_k}h_\xi(u)\,du.
  \label{eq:window-trace-approx}
\end{equation}
Thus the windowed target is the finite-resolution bridge between the state-level definition and the computable time series.
The finite-grid MSPD functional used on a window-indexed trace is defined as follows.  Let \(a=(a_1,\ldots,a_K)\in\mathbb R^K\).  Let \(\mathcal R=\{r_1,\ldots,r_J\}\) be a finite ordered scale grid, and write \(r_j^+=r_{j+1}\).  For each \(r_j\), let
\[
  \mathcal C_{r_j}:\mathbb R^K\to\mathbb R^{K_j}
\]
be a discrete temporal coarse-graining operator, and let
\[
  U_{r_j}:\mathbb R^{K_{j+1}}\to\mathbb R^{K_j}
\]
map the trace coarse-grained at the next scale back to the grid of \(\mathcal C_{r_j}a\).  Block averaging with repetition upsampling is one implementation; the definition only requires fixed linear coarse-graining and reconstruction operators.  The finite-grid MSPD functional is
\begin{equation}
\mathfrak M_{\mathcal R}(a)
  =
  \frac{
    \sum_{j=1}^{J-1}w_j
    \frac{
      \|\mathcal C_{r_j}a-U_{r_j}\mathcal C_{r_j^+}a\|_2^2
    }{
      \|\mathcal C_{r_j}a\|_2^2+\eta_{\mathrm{MSPD}}^2
    }
  }{
    \sum_{j=1}^{J-1}w_j
  }.
  \label{eq:finite-grid-functional}
\end{equation}
The windowed finite-resolution MSPD target is then
\begin{equation}
\operatorname{MSPD}^{\mathrm{win}}_{\xi,\tau}
  =
  \mathfrak M_{\mathcal R}
  (h_{1,\tau}^{\star},\ldots,h_{K,\tau}^{\star}).
  \label{eq:windowed-mspd-target}
\end{equation}
This is the population quantity targeted by the finite computation.

\subsubsection*{3.6 Empirical estimator and consistency statement}
\label{method:finite-estimator}

The empirical estimator replaces each population law \(L_{i,k,\tau}^{\xi}\) by a finite empirical law.  For a sampled local carrier \(i\) in window \(I_k\), choose sample times \(u_{k,1},\ldots,u_{k,m}\in I_{k,\tau}\) and define
\begin{equation}
\widehat L_{i,k,\tau}
  =
  \frac{1}{m}
  \sum_{q=1}^{m}
  \delta_{\frac{Z_i^\xi(u_{k,q}+\tau)-Z_i^\xi(u_{k,q})}{\tau}}.
  \label{eq:empirical-window-law}
\end{equation}
For transition-pair systems the atom in Eq.~\eqref{eq:empirical-window-law} is replaced by \((Z_i^\xi(u_{k,q}),Z_i^\xi(u_{k,q}+\tau))\).  If \(i_1,\ldots,i_n\) are sampled local carriers in window \(I_k\), the empirical pairwise heterogeneity is
\begin{equation}
\widehat H_k
  =
  \frac{2}{n(n-1)}
  \sum_{1\le a<b\le n}
  D\left(
    \widehat L_{i_a,k,\tau},
    \widehat L_{i_b,k,\tau}
  \right).
  \label{eq:empirical-window-H}
\end{equation}
A pooled-null baseline \(\widehat H_k^0\) is computed from pseudo-local empirical laws sampled from the pooled transition samples in the same window.  It corrects the finite-sample distance induced by empirical transition laws when local carrier identity carries no information beyond the pooled transition distribution.  The null-corrected empirical heterogeneity is
\begin{equation}
\widehat{\Delta H}_k
  =
  \widehat H_k-\widehat H_k^0.
  \label{eq:delta-H-estimator}
\end{equation}
The pooled null is an estimator-level correction and is not part of the population MSPD functional.

The empirical trace and empirical MSPD score are
\begin{equation}
\widehat h_{k,\tau}
  =
  \phi(\widehat{\Delta H}_k),
  \qquad
  \widehat{\operatorname{MSPD}}_{\tau}
  =
  \mathfrak M_{\mathcal R}
  (\widehat h_{1,\tau},\ldots,\widehat h_{K,\tau}).
  \label{eq:empirical-mspd}
\end{equation}

Two consistency statements justify the empirical score as an approximation to the pathwise MSPD.  Under standard regularity conditions on the sampled carriers, the empirical transition laws, the pooled-null baseline, and the preprocessing map \(\phi\), the empirical estimator converges in probability to the windowed target:
\begin{equation}
\widehat{\operatorname{MSPD}}_{\tau}
  \xrightarrow[]{p}
  \operatorname{MSPD}^{\mathrm{win}}_{\xi,\tau}
  \label{eq:empirical-to-windowed-consistency}
\end{equation}
(Proposition~1).  Under a local-stationarity condition on the windowed traces and refinement of the scale grid and coarse-graining operators to their continuous counterparts, the windowed target converges to the pathwise MSPD:
\begin{equation}
\operatorname{MSPD}^{\mathrm{win}}_{\xi,\tau}
  \longrightarrow
  \operatorname{MSPD}[h_\xi]
  \label{eq:windowed-to-pathwise-consistency}
\end{equation}
(Proposition~2).  The full statement of both propositions, the precise regularity conditions, and the proofs are given in Appendix~\ref{method:appendix-proof}.

\subsubsection*{3.7 Scope}
\label{method:scope}

MSPD measures the temporal multiscale organization of heterogeneity in local transition laws.  It is defined after specifying the configuration process, the sampling measure over local degrees of freedom, the distance between transition laws, and the evaluation protocol.  These choices determine the dynamical observable whose complexity is measured.  The method does not assign a universal complexity to arbitrary processes; it measures one aspect of dynamical complexity for systems in which local transition laws are meaningful objects of analysis.

\subsection{4 Experiments}
\label{experiments}

This section evaluates MSPD as an empirical instrument for measuring temporally organized heterogeneity in local transition dynamics. The experiments are designed to separate this target from two simpler alternatives: visual novelty and raw motion magnitude. We first specify the substrates and shared evaluation protocol, then report each claim in a self-contained form, with the tested object, control construction, statistic, result, and interpretation stated locally.

\begin{table*}[t]
\centering
\scriptsize
\setlength{\tabcolsep}{3pt}
\renewcommand{\arraystretch}{1.18}
\begin{tabular}{@{}p{0.045\textwidth}p{0.25\textwidth}p{0.25\textwidth}p{0.28\textwidth}p{0.08\textwidth}@{}}
\toprule
Claim & Scientific claim & Operational test & Reported statistic & Figure \\
\midrule
N0
& MSPD should reproduce the obvious complexity ordering on basic synthetic examples: simulations with minimal organized spatial or temporal structure should receive lower scores than simulations with more structured spatial or temporal dynamics.
& We evaluate MSPD on synthetic families selected to cover three basic regimes: low-complexity negative controls, spatial organization through persistent local roles, and temporal organization where the relevant structure changes over time.
& Family-level MSPD, processed $\Delta H$, event localization, and role-recovery scores against the known synthetic labels.
& Figs.~\ref{fig:synthetic-summary}, \ref{fig:synthetic-heatmaps} \\
\addlinespace[0.60em]

C1
& MSPD-optimized parameters produce higher post-hoc complexity scores than matched random parameters from the same substrate.
& For each substrate, we compare optimized checkpoints with random checkpoints sampled from the same parameterization, using the same substrate-specific post-hoc MSPD reporting protocol.
& $\Delta^{C1}_{s,r}$ is the optimized score minus the median matched-random score, $S_s(\theta^{\mathrm{opt}}_{s,r})-\mathrm{median}_{j}S_s(\theta^{\mathrm{rand}}_{s,r,j})$.
& Figs.~\ref{fig:ca-c1}, \ref{fig:c1-flow}, \ref{fig:c1-plife}, \ref{fig:c1-tau-profiles} \\
\addlinespace[0.60em]

C2
& High-$\Delta H$ states should correspond to states with greater divergence among controlled alternative futures.
& Saved branch states are assigned a present-time branch energy $E_b$ and resumed under a substrate-appropriate continuation family while the rule remains fixed.
& The relationship between $E_b$ and future divergence $B_b$: a continuous association for CA and within-run high-minus-low contrasts for Flow-Lenia.
& Figs.~\ref{fig:ca-c2}, \ref{fig:c2-flow} \\
\addlinespace[0.60em]

C5
& Higher MSPD corresponds to stronger scale-dependent frustration: high-complexity systems exhibit larger differences between the dynamics expressed at different spatial extents.
& For each checkpoint, we compare a free continuation with a matched continuation subjected to an early spatial constraint, then subtract ordinary free-future variability measured under the same substrate-specific continuation law.
& $\Delta^{C5}_{s,r}$ is the optimized intervention score minus the median matched-random intervention score; $F_s$ subtracts ordinary future variability from the intervention effect.
& Figs.~\ref{fig:c5-flow}, \ref{fig:c5-plife} \\
\bottomrule
\end{tabular}
\caption{Claim-level organization of Section~4. The table separates each scientific claim from the operational test used to evaluate it. C1 and C5 use independent matched optimization groups as statistical units. C2 uses branch states within source trajectories, with Flow-Lenia inference aggregated at the independent optimization-run level. Scale separation, biodiversity, and NN-OEE transfer are not claims in this section.}
\label{tab:claim-map}
\end{table*}

\FloatBarrier

\paragraph{Note on claim numbering.}
Two additional claims from the pre-registered design are
intentionally absent from this section.
C3 stated that the temporal scale selected by MSPD corresponds
to the scale at which stable, spatially coherent dynamical
roles are most clearly separated; this claim was not
experimentally confirmed under the current protocol.
C4 stated that optimizing MSPD produces non-degenerate dynamics
with ecological or phenotypic richness comparable to systems
optimized by neural-network-based open-ended-evolution
objectives; we did not find a measurable operationalization of
this comparison within the present scope.
We keep the original numbering (C1, C2, C5) to preserve
consistency with the pre-registered claim table.  If reviewers
prefer a contiguous sequence, we are happy to renumber
C5$\to$C3 in a revision.

\subsubsection*{4.1 Substrates, optimized objects and local transition laws}
\label{experiments:substrates}

The same MSPD construction is applied to different substrates by changing only the local carrier and the empirical local transition law.  Table~\ref{tab:optimized-objects} gives the object optimized or selected in each substrate; the paragraphs below define the observable used by the estimator and the optimization object used in C1.  Appendix~C gives the exact optimization, random-control and post-hoc evaluation protocols.

\begin{table}[H]
\centering
\footnotesize
\begin{tabular}{p{0.16\linewidth}p{0.30\linewidth}p{0.43\linewidth}}
\toprule
Substrate & Optimized or selected object $\theta$ & Evaluation object used by MSPD \\
\midrule
Life-like CA & 18-bit totalistic birth/survival rule & Categorical law of local before-after neighborhood events \\
Flow-Lenia & Rule coordinates evaluated together with a fixed realised pathwise context & Empirical laws of finite-time displacements of passive mass probes \\
Particle Life++ & Neural pair-interaction weights in $f_\theta(c_i,c_j)$ & Empirical laws of finite-time particle displacements \\
\bottomrule
\end{tabular}
\caption{Substrate-specific measurement objects.  The optimized parameter is not always the same kind of object: it is a discrete rule for CA, Flow-Lenia rule coordinates evaluated in a fixed realised context, and neural interaction weights for Particle Life++.  The reported MSPD score is always computed post-hoc from empirical local transition laws.}
\label{tab:optimized-objects}
\end{table}
\FloatBarrier

\paragraph{Life-like cellular automata.}
A Life-like cellular automaton is simulated on a periodic lattice $\Lambda$ with alphabet $A=\{0,1\}$ and Moore neighborhood $B=\{-1,0,1\}^2$:
\[
  x_t\in A^\Lambda,
  \qquad x_{t+1}(i)=f_\theta(x_t|_{i+B}).
\]
The selected object $\theta$ is the 18-bit totalistic rule code.  For C1, the optimization protocol is an exhaustive sweep over the same $2^{18}$ Life-like rule space; the reported optimized rule is the top MSPD rule from this sweep, and matched controls are random rules drawn from the same space.  Since there is no velocity variable, the local transition event is categorical,
\[
  a_t(i)=\bigl(x_t|_{i+B},x_{t+1}(i)\bigr)\in A^B\times A,
\]
and the empirical local transition law in window $I_k$ is
\[
  \widehat T_i(I_k)(a)=\frac{1}{|I_k|}\sum_{t\in I_k}\mathbf 1[a_t(i)=a].
\]
Only sites active within the local neighborhood during the window are weighted in sparse boards.  This prevents empty background from dominating the transition-law average. In the MSPD notation, this corresponds to choosing the spatial averaging measure $\mu_w$ as the normalized counting measure on the window-active support,
\[
A_w=\{x:\text{the local neighborhood of }x\text{ is active at least once during window }w\},
\qquad
\mu_w(x)=\frac{\mathbf 1{x\in A_w}}{|A_w|}.
\]
Thus the reported MSPD average is not taken with respect to the uniform measure over the full board, but with respect to the empirical active-support measure induced by the current window.

\paragraph{Flow-Lenia.}
Flow-Lenia is a continuous cellular automaton on a wall-bounded grid.  Its state consists of a transported mass field $A_t:\Omega\to\mathbb R_{\ge 0}^{C}$ and a carried internal field $P_t:\Omega\to\mathbb R^{K}$, and one update
\[
  (A_{t+1},P_{t+1},F_t)=\Phi_\phi(A_t,P_t)
\]
is produced by convolutional response, pointwise growth/gating, and RT reintegration of the mass and internal fields through a rule-induced transport field $F_t$, with categorical internal-state mixing and stochastic mutation.  The full update composition and simulator parameters are given in Appendix~\ref{app:flow-lenia-algorithm}.  Initial conditions are constructed from random local patches and advanced by one simulator step before use, so exact replay fixes the returned post-step state.

For the experiments, a realised Flow-Lenia checkpoint decomposes as
\[
  \theta=(\phi,\chi),
\]
where $\phi$ is the optimizer-updated rule component and $\chi$ is the fixed pathwise context: the returned initial $A/P$ state, the stochastic key realization and split schedule, and the simulator/execution configuration required to reproduce the path.  Only $\phi$ is optimized; each independent optimization run evaluates candidate rules on a fixed bank of realised contexts, and C1 replays those contexts exactly.  The search vector also contains a nuisance coordinate that selects the MSPD lag; Appendix~C gives the full optimization protocol.

MSPD is evaluated on Lagrangian tracer trajectories $q_a(t)$ advanced after each Flow-Lenia step by the same RT reintegration-tracking machinery as the substrate (Appendix~\ref{app:flow-lenia-algorithm}).  The saved coordinates define the local empirical law used by MSPD,
\[
  \widehat V_a(I_k,\tau)
  =
  \frac{1}{|I_{k,\tau}|}
  \sum_{t\in I_{k,\tau}}
  \delta_{\Delta_\Omega(q_a(t),q_a(t+\tau))},
\]
so Flow-Lenia MSPD measures heterogeneity in finite-lag transported-mass motion.

A pathwise stochasticity audit (Appendix~\ref{app:flow-stochasticity-audit}) shows that within-rule continuations from bit-exact shared states, differing only in the simulator's internal stochastic realization, rapidly diverge to the same order as between-rule comparisons.  Changing only the stochastic realization can therefore alter the Flow-Lenia path as strongly as changing the rule.  We consequently treat the realised stochastic context as part of the fixed pathwise checkpoint rather than averaging over it, which the available optimization budget would not permit.

\paragraph{Particle Life++.}
Particle Life++ is a neural pair-interaction particle system on a toroidal substrate with state $S_n=\{(x_i^n,v_i^n,c_i^n)\}_{i=1}^{N}$ (position, velocity and unit-norm color).  For each ordered pair $(i,j)$, a small neural network $(\alpha_{ij},g_{ij})=f_\theta(c_i,c_j)$ maps colors to an interaction strength and a color update.  These pair outputs accumulate over neighbors into a total force $F_i$ and color derivative $\dot c_i$, which are then integrated with damping to yield $v_i^{n+1}$, $x_i^{n+1}$ (with toroidal wrap) and a renormalized $c_i^{n+1}$.  The full pair geometry, radial kernel and integration step are given in Appendix~\ref{app:plife-plus-algorithm}.  Particle identities are the local carriers, MSPD uses empirical laws of finite-time particle displacements, and the optimized object is the neural weight vector $\theta$ in $f_\theta$.

Random neural weights in this substrate often settle into low-motion stationary basins where forces cancel or are damped before sustained motion develops, giving narrow and similar displacement laws across windows and suppressing both mean heterogeneity and MSPD.  $\theta$ is optimized by Sep-CMA-ES with a mean-heterogeneity term added to leave this basin before the MSPD term selects temporally structured regimes.  The reported score is a post-hoc MSPD score computed by the same protocol for optimized checkpoints and matched random controls.

\subsubsection*{4.2 Shared evaluation choices}
\label{experiments:shared}

The experiments use matched controls because MSPD scales depend on the substrate, rollout budget and local transition representation.  For substrate $s$ and independent group $r$, the optimized checkpoint is $\theta^{\mathrm{opt}}_{s,r}$ and the matched random controls are
\[
  \theta^{\mathrm{rand}}_{s,r,1},\ldots,\theta^{\mathrm{rand}}_{s,r,J}.
\]
A matched random control is sampled from the same optimization initialization distribution used by the optimizer but is not selected by the optimization objective. The comparison therefore asks whether the optimization procedure found a higher-MSPD region of the same parameter space, rather than whether one substrate has larger raw MSPD than another.

For Flow-Lenia and Particle Life++, the optimization-time estimator is lower budget than the reported estimator.  All C1 claims use a fixed post-hoc protocol:
\[
  S(\theta)=\widehat{\operatorname{MSPD}}(\theta).
\]
The two continuous substrates retain interleaved temporal partitions so that diagnostic and reporting windows cover comparable rollout phases.  Particle Life++ uses its existing post-hoc selection/evaluation split.  In Flow-Lenia, the lag is inherited from the optimization-selected candidate and fixed for the optimized rule and all matched random controls in that group.  Even-indexed split A is diagnostic and odd-indexed split B is the reporting partition.  Split B is temporally disjoint from split A, but it is not an independent fresh-seed holdout from optimization because the selected candidate was chosen from the complete fixed pathwise objective.

C2 and C5 require distances between future rollouts. Direct pixel-space distances are a poor default for this purpose because they compare synchronized arrays rather than simulation regimes.  A coherent object translated from one corner of the frame to another can dominate pixel MSE even when the qualitative regime is unchanged; conversely, two independent gaussian noise renderings can have large pixel-wise discrepancy while representing the same visually homogeneous regime.  This is the same failure mode that motivates learned perceptual distances in image comparison: deep visual features have been shown to correlate better with human perceptual similarity than shallow pixel-level metrics such as PSNR or SSIM~\cite{zhang2018perceptual}, and more recent metrics explicitly target holistic human similarity judgments beyond local pixel or patch agreement~\cite{fu2023dreamsim}.

We therefore use a fixed image encoder as a pragmatic representation-space readout for future-rollout separation, following the foundation-model evaluation practice of ASAL~\cite{kumar2024automating}.  The encoder is not used to define MSPD and is not part of the optimization objective.  It is used only after rollouts have been generated, to quantify whether two future continuations occupy similar or different rendered outcome regimes.  The exact encoder backend and preprocessing are specified in Appendix~\ref{app:encoder-description}.  For a rendered frame $R_a(t)$ from rollout $a$, we compute the normalized feature
\[
  z_a(t)=\frac{f(R_a(t))}{\|f(R_a(t))\|_2},
  \qquad
  Z_a=\{z_a(t):t\in U_a\}.
\]
With cosine distance $d_{\cos}(u,v)=1-u^\top v$, two rollout futures are compared as feature clouds by the symmetric Chamfer distance
\[
  d_{\mathrm{Ch}}(Z_a,Z_b)=
  \frac{1}{2}\left(
  \frac{1}{|Z_a|}\sum_{u\in Z_a}\min_{v\in Z_b}d_{\cos}(u,v)
  +
  \frac{1}{|Z_b|}\sum_{v\in Z_b}\min_{u\in Z_a}d_{\cos}(u,v)
  \right).
\]

This choice should be read as a measurement proxy, not as a ground-truth notion of dynamical equivalence.  Neural image encoders are trained on particular image distributions, and ALife renderings---especially cellular automata or sparse particle systems---can be out of domain.  In such cases the feature distance may be noisier or may emphasize visual attributes that are not dynamically central.  We nevertheless use this readout because C2 and C5 require a phase-tolerant distance between future rendered outcomes, and no pixel-space alternative provides a reliable comparison of translated, phase-shifted, or visually noisy simulations at the qualitative level relevant for these tests.

\subsubsection*{4.3 N0: controlled synthetic calibration}
\label{experiments:synthetic}

\textbf{Purpose.}  N0 checks that the finite estimator responds to transition-law structure whose ground truth is known.  The test separates raw motion, stationary role heterogeneity, temporally extended transitions and split events before the estimator is applied to optimized ALife systems.

Each synthetic rollout contains particles on the torus,
\[
  q_i(t)\in \mathbb T^2=[0,1)^2,
  \qquad i=1,\ldots,N,
\]
with shortest-periodic displacement
\[
  \Delta_{\mathbb T^2}(q,q')=((q'-q+1/2)\bmod 1)-1/2.
\]
For a window $I_k$ and lag $\tau$, the empirical transition law is
\[
  \widehat V_{i,k,\tau}
  =\frac{1}{|I_{k,\tau}|}
  \sum_{t\in I_{k,\tau}}
  \delta_{\Delta_{\mathbb T^2}(q_i(t),q_i(t+\tau))},
  \qquad I_{k,\tau}=\{t\in I_k:t+\tau\in I_k\}.
\]
MSPD is computed from the resulting windowed heterogeneity trace.

Rendered examples of the synthetic calibration families are provided in Appendix~\ref{app:synthetic-rendered-examples}; the N0 scores themselves are computed from particle trajectories rather than from rendered frames.

\begin{table}[H]
\centering
\scriptsize
\begin{tabular}{p{0.07\linewidth}p{0.31\linewidth}p{0.29\linewidth}p{0.20\linewidth}}
\toprule
Family & Generator & Interpretation target & Observed response \\
\midrule
S0 & Static particles & zero-motion null & MSPD $=0$ \\
S1 & Homogeneous Brownian or Gaussian motion & motion without local role structure & low MSPD $3.46\times10^{-5}$ \\
S3 & One coherent moving blob & organized but locally homogeneous motion & near-zero MSPD $5.45\times10^{-8}$ \\
S4 & Two fixed role velocities & stationary heterogeneity without temporal organization & high amplitude, low MSPD \\
S5 & Global synchronous velocity switch & changepoint without role spread & low event-localization signal \\
S6 & Staggered switch times & temporally extended heterogeneity & event error $0$, high MSPD \\
S7 & Multi-scale moving blobs & scale calibration & selected lag in expected range \\
S8 & Blob splitting into moving groups & split event plus role emergence & event error $0$, largest MSPD \\
\bottomrule
\end{tabular}
\caption{Synthetic calibration families.  The suite includes null controls, stationary heterogeneity controls and time-localized transition generators, so a high score cannot be attributed merely to motion or to a constant mixture of roles.}
\label{tab:synthetic-calibration}
\end{table}

\begin{figure}[H]
\centering
\includegraphics[width=0.92\linewidth]{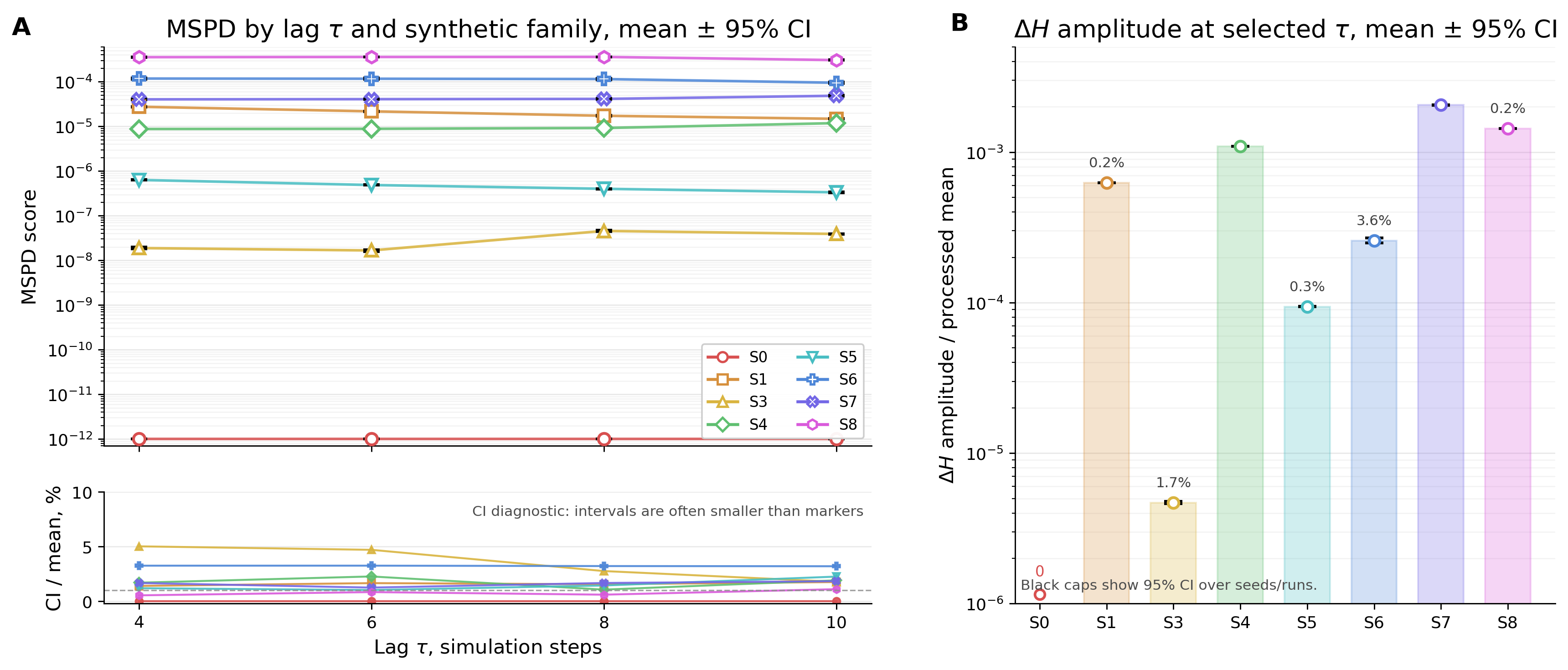}
\caption{Synthetic calibration summary.  Panel A reports selected MSPD scores by family; Panel B reports processed $\Delta H$ amplitude.  The high-MSPD cases are the temporally extended switch and split-transition generators.}
\label{fig:synthetic-summary}
\end{figure}
\FloatBarrier

\begin{figure}[H]
\centering
\includegraphics[width=0.92\linewidth]{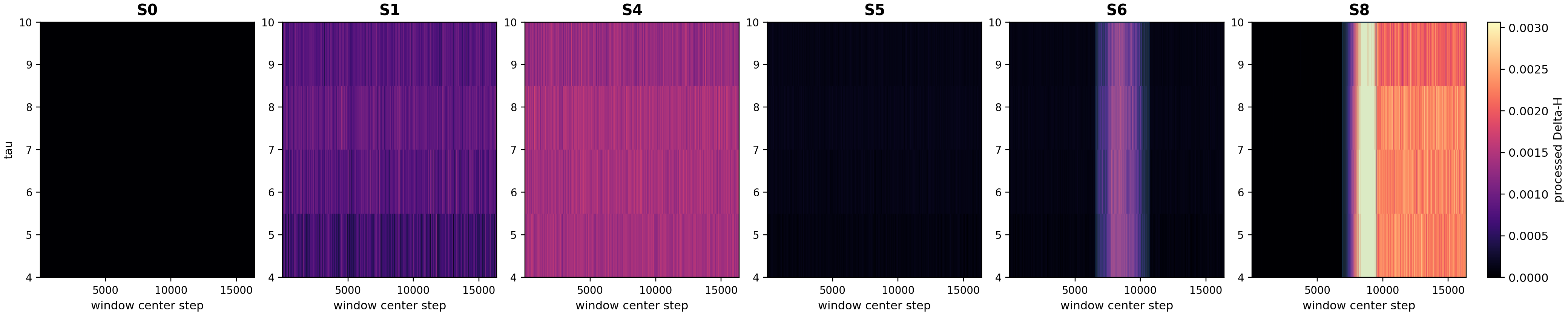}
\caption{Synthetic $\Delta H$ maps.  Horizontal position is window index, vertical position is lag $\tau$, and color is local transition heterogeneity.  Temporally extended transition systems concentrate high $\Delta H$ around designed transition intervals, while null and homogeneous controls remain low.}
\label{fig:synthetic-heatmaps}
\end{figure}
\FloatBarrier

\textbf{Result and interpretation.}
The synthetic results separate three quantities that can otherwise be confused: visible motion, the average level of local heterogeneity measured by processed $\Delta H$, and MSPD as temporal organization of that heterogeneity trace.  The zero-motion null S0 and the homogeneous-motion controls S1 and S3 give near-zero $\Delta H$ and MSPD, confirming that neither stochastic motion nor coherent object motion is by itself complexity for MSPD.  The stationary role-heterogeneity control S4 gives high $\Delta H$ amplitude but low MSPD, separating mean heterogeneity from temporally organized complexity.  The synchronous global switch S5 produces a low MSPD response because the change is shared across all particles at once and is not resolved by the chosen window and lag grid; this is a resolution-dependent limitation whose remedy is finer temporal logging.  The temporally extended transition families S6, S7 and S8 give the largest MSPD scores, and the split-with-role-emergence family S8 gives the largest MSPD of all even though its mean $\Delta H$ is below S7, illustrating that MSPD rewards temporally organized changes rather than the largest average heterogeneity.  Full per-family analysis is given in Appendix~\ref{app:synthetic-calibration-protocol}.

Taken together, N0 shows that the finite estimator behaves as intended on basic cases.  MSPD remains low for static dynamics, homogeneous motion, coherent common motion, and mostly global changes.  It increases when heterogeneity is organized across time through staggered transitions, split events, or evolving role structure.  In this sense, ``complexity'' in the reported MSPD score means multiscale temporal organization of local transition-law heterogeneity, not raw motion magnitude, visual novelty, or the average $\Delta H$ level alone.

\subsubsection*{4.4 C1: optimization against matched random controls}
\label{experiments:c1}

\textbf{Purpose.}  C1 tests whether optimizing the MSPD objective produces checkpoints with larger post-hoc MSPD than matched random controls from the same substrate parameterization.  The comparison is within substrate and within matched group.

\textbf{Statistic.}  For substrate $s$ and independent group $r$,
\[
  \Delta^{C1}_{s,r}
  =S_s(\theta^{\mathrm{opt}}_{s,r})
  -\operatorname{median}_{j}S_s(\theta^{\mathrm{rand}}_{s,r,j}).
\]
The reported evidence is the number of positive matched groups, the median contrast and an exact one-sided sign test over matched groups.  The sign test is used because the number of independent optimization groups is small and no normal approximation is required.

\begin{table}[H]
\centering
\footnotesize
\begin{tabular}{p{0.20\linewidth}p{0.20\linewidth}p{0.20\linewidth}p{0.16\linewidth}p{0.16\linewidth}}
\toprule
Substrate & Positive groups & Median contrast & Sign-test $p$ & Status \\
\midrule
Life-like CA & separated optimized set & large positive shift & not applicable & confirmed by sweep \\
Flow-Lenia & $9/10$ & $4.69\times10^{-4}$ & $0.0107$ & supported for fixed objectives \\
Particle Life++ & $7/7$ & $0.00267$ & $0.0078$ & confirmed \\
\bottomrule
\end{tabular}
\caption{C1 results.  The CA result is a discrete rule-search result; Flow-Lenia and Particle Life++ are matched continuous-parameter optimization results.}
\label{tab:c1-results}
\end{table}
\FloatBarrier

All three substrates show positive matched contrasts: the CA sweep separates selected from random rules; Flow-Lenia is positive in $9/10$ optimization runs on the fixed pathwise objectives under exact optimizer-native replay (with Flow-Lenia support restricted to those fixed objectives and inheriting the optimization-selected lag for both the optimized and random members of each matched group); Particle Life++ is positive in all $7/7$ groups.  Substrate-specific narratives, figures, and the lag/temporal-split audit are given in Appendix~\ref{app:c1-substrate-results}.
\FloatBarrier

\subsubsection*{4.5 C2: local heterogeneity and future outcome divergence}
\label{experiments:c2}

\textbf{Purpose.}  C2 tests a state-level interpretation of $\Delta H$.  It asks whether a saved state with larger present-time local transition heterogeneity has a broader set of controlled alternative futures.

\textbf{Design.}  A branch state $b=(i,t)$ is a saved state from source trajectory $i$ at branch time $t$.  Its present-time score is the branch energy
\[
  E_b
  =\frac{1}{|\mathcal T_{\mathrm{adm}}|}
  \sum_{\tau\in\mathcal T_{\mathrm{adm}}}\phi(\Delta H(w_b,\tau)),
\]
where $w_b$ is the window containing $b$, $\mathcal T_{\mathrm{adm}}$ is the admissible lag set, and $\phi$ is the nonnegative preprocessing used by the MSPD pipeline.  The same saved state is used to generate $M$ controlled alternative continuations under the same rule,
\[
  Y_{b,1},\ldots,Y_{b,M}.
\]
The continuation family is substrate-specific: deterministic CA uses small board perturbations, whereas stochastic Flow-Lenia preserves the exact state and samples independent realizations of the simulator's internal stochasticity.  Future divergence is
\[
  B_b=\operatorname{median}_{p<q}d_{\mathrm{out}}(Y_{b,p},Y_{b,q}).
\]
The branch sampler covers low, middle and high quantiles of $E_b$ within each source trajectory.  These labels are sampling strata rather than additional claims.

\textbf{Statistic.}  CA reports the continuous association between $E_b$ and $B_b$.  Flow-Lenia uses the stratified design to form high-minus-low future-divergence contrasts within each source run and aggregates inference over independent optimization runs.  This keeps the scientific question fixed while respecting the different deterministic and stochastic continuation laws of the two substrates.

CA supplies a discrete-substrate check via bit-flip perturbations and Hamming future divergence; Flow-Lenia supplies the quantitative test via exact-state stochastic continuations, where all $10/10$ independent optimized source runs show positive high-minus-low future-divergence contrasts after aggregation over the reported horizons.  Substrate-specific continuation mechanisms, figures, and the interpretation-and-boundary discussion are given in Appendix~\ref{app:c2-substrate-results}.
\FloatBarrier

\subsubsection*{4.6 C5: blockwise frustration and history dependence}
\label{experiments:c5}

\textbf{Purpose.}  C5 tests whether optimized systems depend on early spatial co-development.  The intervention is designed to disrupt early spatial integration while keeping the checkpoint $\theta$ fixed.  Late futures are compared after the intervention has been released, so the measurement is not merely the immediate damage caused by inserting walls or shuffling cells.

\textbf{Design.}  For each checkpoint $\theta$, an intervention continuation is paired with a free continuation from the same saved state and matched continuation randomness.  Additional free continuations estimate ordinary future variability.  The intervention implementation is substrate-specific: Flow-Lenia uses temporary isolated spatial blocks, whereas Particle Life++ uses a cell-shuffle intervention after warmup.  Both implementations instantiate the same abstract intervention: disrupt large-scale spatial co-organization without changing $\theta$.

\textbf{Statistic.}  Write $I_s(\theta)$ for the post-release intervention-versus-free distance and $V_s(\theta)$ for ordinary free-future variability.  The checkpoint-level anchored frustration score is
\[
  F_s(\theta)=I_s(\theta)-V_s(\theta).
\]
The first term measures the effect of the early intervention; the second subtracts ordinary continuation variability for the same checkpoint.  The matched optimization contrast is
\[
  \Delta^{C5}_{s,r}
  =F_s(\theta^{\mathrm{opt}}_{s,r})
  -\operatorname{median}_{j}F_s(\theta^{\mathrm{rand}}_{s,r,j}).
\]
A positive value means that optimized checkpoints show larger intervention-minus-seed effects than matched random controls.

\begin{table}[H]
\centering
\footnotesize
\begin{tabular}{p{0.20\linewidth}p{0.30\linewidth}p{0.14\linewidth}p{0.16\linewidth}p{0.12\linewidth}}
\toprule
Substrate & Intervention & Positive groups & Median contrast & Sign-test $p$ \\
\midrule
Flow-Lenia & temporary mass-preserving blocks, then release & $8/10$ & $5.83\times10^{-4}$ & $0.0547$ \\
Particle Life++ & warmup, $2\times2$ cell shuffle, then release & $6/7$ & $0.00477$ & $0.0625$ \\
\bottomrule
\end{tabular}
\caption{C5 results.  Both substrates show positive matched contrasts in most independent groups, but the current group counts support a directional history-dependence result rather than a high-confidence significance claim.}
\label{tab:c5-results}
\end{table}
\FloatBarrier

Flow-Lenia uses temporary $3\times3$ compartmentalization with a mass-preserving projection during the wall phase, followed by release; Particle Life++ uses a warmup-then-cell-shuffle intervention.  Both substrates show positive matched contrasts in most independent groups ($8/10$ for Flow-Lenia at the $5$k sensitivity horizon; $6/7$ for Particle Life++), supporting a directional history-dependence result under the specified interventions and temporal scales.  Substrate-specific narratives and figures are given in Appendix~\ref{app:c5-substrate-results}.
\FloatBarrier

\subsubsection*{4.7 Section-level conclusion}
\label{experiments:conclusion}

The experiments support a narrow but coherent interpretation.  The synthetic suite validates that MSPD responds to temporally organized local heterogeneity.  C1 shows that the score is optimizable in all three substrates under their reported protocols, with Flow-Lenia support restricted to the fixed pathwise objectives used during search.  C2 shows that high-$\Delta H$ Flow-Lenia states have more separated intrinsic stochastic futures and that the same state-level logic transfers to deterministic CA through controlled state perturbations.  C5 shows directionally consistent history-dependence effects in Flow-Lenia and Particle Life++, tied to the specified interventions and temporal scales.

\begin{center}\rule{0.5\linewidth}{0.5pt}\end{center}

\subsection{5 Discussion}\label{discussion}

% discussion: [C5] -- interprets the scale-dependent frustration finding via spin-glass framing
\textbf{5.1 Scale-dependent frustration as a physical signature.}
The C5 finding --- that MSPD-optimised systems exhibit a larger
intervention-based gap between constrained and unconstrained dynamics
than matched random controls --- gives the frustration / non-ergodicity
criterion of Vanchurin et al.~\cite{vanchurin2022toward} a directly
measurable operationalisation in ALife substrates. In the spin-glass
framing of Laughlin et al.~\cite{laughlin1,laughlin2} and Wolf et al.~\cite{wolf2018physical}, frustration manifests
as competing local minima whose effective dynamics depend on the
spatial extent over which the system is allowed to relax. The
constrained-rollout protocol exposes exactly this: under an early
spatial constraint, optimised systems explore a noticeably different
region of state space than they would unconstrained, while random
controls do not. MSPD thus picks up frustration as a downstream
consequence of selecting for multiscale dynamic organization, rather
than as an explicitly optimised target.

% discussion: [C5] -- design implication: MSPD as one principled driver among possible objectives
\textbf{5.2 Implications for OEE objective design.}
A practical consequence of C5 is that MSPD provides a single, explicit
fitness function that produces systems satisfying physical life
criteria (including frustration) without being asked to. This is the property
that distinguishes MSPD from black-box drivers such as
ASAL~\cite{kumar2024automating} or
Leniabreeder~\cite{faldor2024leniabreeder}: the latter expand the
space of \emph{patterns} a simulation produces via learned
embeddings, but the connection to physical theories of complexity
remains implicit. Multi-objective evolution that combines MSPD with a
complementary snapshot-structural metric is a natural next step and
may dominate either driver alone, but we leave that comparison to
future work.

% discussion: cross-substrate transfer -- internal DoF as precondition for the local-transition-law decomposition
\textbf{5.3 ALife substrates and internal degrees of freedom.}
The cross-substrate transfer of C1/C2/C5 requires only that the
substrate exposes meaningful local transition laws --- either as
categorical rule events (Life-CA) or as displacement distributions
(Flow-Lenia, Particle Life++). Substrates whose local transition
laws collapse to a single nearly-deterministic atom across all
sites, such as Particle Life++ in the zero-motion basin, give MSPD
near zero by construction; this is the expected behaviour and gives
the metric a built-in null. The decomposition is most informative
in substrates with internal degrees of freedom (state, type,
memory) that let local transition laws differ meaningfully between
carriers.

% discussion: [C2] + cross-substrate transfer -- conceptual note on when ΔH does and does not predict branch instability
\textbf{5.4 What $\Delta H$ does and does not signal.}
A locally elevated $\Delta H$ --- high relative to neighbouring
windows of the same rollout --- is best read as a
\emph{detector} that the system is passing through a change
in its dynamics, not as a criterion that pins down which
change is occurring. The detector logic is straightforward:
when a rollout crosses a regime boundary, some carriers will
typically have already entered the new dynamical regime while
others still carry the old one, and the within-window
heterogeneity of their local transition laws rises
correspondingly. Perturbing such a state then selects between
competing late futures and yields the C2 signature.

The converse, however, does not hold. Many situations can
push $\Delta H$ above the local baseline without a global
regime change being responsible. One example is rare local
events on a small subset of carriers --- isolated collisions,
brief bursts, transient role switches at finite simulation
horizon --- which raise the within-window heterogeneity even
when the global dynamics are stationary in distribution.
Other mechanisms (boundary or finite-size artifacts,
heterogeneous initial conditions that have not yet relaxed,
substrate-specific intermittent modes) act similarly. We do
not attempt a taxonomy here; the point is only that high
$\Delta H$ is a necessary condition for the C2 association,
not a sufficient one.

This non-equivalence is what we observe on Particle Life++.
The substrate has deep attractors: even strong injected noise
relaxes back toward the pre-perturbation regime within a few
windows, so the attractor structure short-circuits the
exact-state stochastic-continuation channel that C2 now tests:
branches launched from the same saved state under independent
continuation seeds re-converge into the same basin within a
few windows, so high-$\Delta H$ states do not translate into
measurable future divergence. Local $\Delta H$ can still rise
--- the substrate produces transient mixing events that look
like regime changes within a window --- but those rises are
not coupled to branch divergence, because the underlying
dynamics return to the same basin. We therefore
treat C2 transfer to Particle Life++ as outside the present
scope (\S6) and as conceptually distinct from C5 on the same
substrate, which probes intervention dependence at a much
longer timescale than perturbation-recovery.

\begin{center}\rule{0.5\linewidth}{0.5pt}\end{center}

\subsection{6 Limitations}\label{limitations}

\textbf{Within-window stationarity.} The consistency of the
windowed estimator (Propositions~1--2) rests on approximate
stationarity of local transition laws within each window
$I_k$. In regimes with abrupt regime shifts --- nucleation of
a new structure, sudden collapse of a basin --- this assumption
is violated locally, and the empirical $\widehat{H}_k$ mixes
the pre- and post-event statistics. We mitigate this with
short windows relative to the dominant timescale and with the
pooled-null correction, but the residual bias is not
quantified and may attenuate \textbf{C1} and \textbf{C2} on
runs that contain such transitions.

\textbf{Statistical power.} Our matched-random comparisons
for \textbf{C1} and \textbf{C5} use a modest number of
independent optimization runs per substrate (Section~4).
For C1, the fixed-context replay contrast reaches
conventional significance on Flow-Lenia ($9/10$ groups,
$p=0.0107$) and remains detected qualitatively on the other
substrates; fine quantitative claims --- effect sizes,
distributional tails of the matched-random null,
substrate-specific exponents --- are still not supported at
this sample size. For C5 the contrast is directionally
positive on both substrates (Flow-Lenia $8/10$, $p=0.0547$ at
the $5$k sensitivity horizon; Particle Life++ $6/7$,
$p=0.0625$) but neither reaches conventional significance at
the reported horizon; the C5 result should therefore be read
as \emph{underpowered evidence of history dependence} rather
than as a confirmed effect. In both cases
the natural strengthening is replication with a larger
matched-group ensemble, without methodological change.

\textbf{Sensitivity of C5 to the intervention protocol.}
The scale-dependent frustration signature (\textbf{C5}) is
measured under a particular constrained-rollout intervention:
constraint geometry, perturbation magnitude, and release time
are fixed by the protocol described in Section~4. Preliminary
checks suggest the qualitative ordering is preserved across
small variations, but a full sensitivity sweep over the
intervention family is left for follow-up work.

\textbf{Incomplete cross-substrate matrix.} The
cross-substrate transfer panel set is sparse rather than
complete. We report C1 on all three substrates (Flow-Lenia,
Life-CA, Particle Life++), C2 on Flow-Lenia and Life-CA, and
C5 on Flow-Lenia and Particle Life++. Two cells are missing:
C5 on Life-CA --- straightforward to add in a follow-up given
the discrete-substrate intervention is simple --- and C2 on
Particle Life++, which is conceptually obstructed by the
deep-attractor structure of the substrate (\S5.4) rather than
by methodology. The cross-substrate transfer should therefore be
read as \emph{partial}: the MSPD construction applies uniformly across the
three substrates, and each individual claim transfers to at
least one non-primary substrate, but a fully populated
$3\times3$ claim-substrate matrix is left to follow-up work.
Other continuous-Lagrangian substrates (Boids, Particle Lenia,
neural cellular automata) carry local transition laws of the
same form and are natural targets, but have not been evaluated
here. Substrates without meaningful local DOFs (e.g.\
single-particle simulations) fall outside the scope of the
metric by construction.

\textbf{No direct head-to-head with NN-OEE drivers.}
This version of the paper does not compare MSPD-as-driver
against neural-network open-endedness objectives (ASAL,
Leniabreeder) on a shared task. The cleanest formulation ---
correlating MSPD loss with an NN-OEE loss across a matched
ensemble --- is deferred to a follow-up paper; preliminary
runs were inconclusive at current sample size.

\begin{center}\rule{0.5\linewidth}{0.5pt}\end{center}

% supports: [C1] [C2] [C5] -- conclusion recaps all claims retained in this version
\subsection{7 Conclusion}\label{conclusion}

% what + why
We set out to bridge a methodological gap in artificial life:
open-ended evolution is presently driven by uninterpretable,
neural-network complexity metrics that cannot be connected to physical
theories of life, while existing physics-grounded metrics
(spatial MSSC) were limited to static patterns.

% what we did (method recap)
We introduced MSPD (Multi-Scale Path Divergence, $D_P$), a
temporal extension of MSSC that inherits its
renormalization-group architecture intact: the spatial
coarse-graining scale is replaced by an observation window,
and cross-scale dissimilarity is computed in
trajectory-distribution space rather than image space. MSPD is
defined at the population level as a functional of the realised
trajectory via local transition laws, with a windowed
finite-resolution estimator and an explicit consistency
statement (Propositions~1--2). As an explicit scalar it plays a
dual role: as a gradient-free fitness function, and as a
post-hoc analytical lens on any simulation that exposes local
transition laws.

% main claims (C1, C2, C5)
Empirically, MSPD-optimized parameters produce higher scores
than matched random parameters from the same substrate under
fixed-context replay of the exact pathwise objective
(\textbf{C1}; fresh-seed and fresh-initial-state
generalization is left to follow-up), and high-$\Delta H$
states along optimized trajectories yield larger future
divergence under exact-state stochastic continuations than
matched low-$\Delta H$ states (\textbf{C2}) --- the metric
responds to the intrinsic dynamics of the sampled trajectories
rather than to injected noise. Used as a lens, MSPD reveals
that optimised systems also exhibit elevated
\emph{scale-dependent frustration} under matched
constrained-rollout interventions (\textbf{C5}) --- linking
MSPD post-hoc to the frustration criterion of Vanchurin et
al.~\cite{vanchurin2022toward} and to the spin-glass framing of
biological complexity of Wolf et al.~\cite{wolf2018physical},
without the metric being optimised for frustration at any
stage. The same protocol transfers to qualitatively different
substrates: C1, C2 and C5 also hold on Life-like cellular
automata and Particle Life++.
% [C1] [C2] [C5]

A single explicit-formula scalar therefore both \emph{directs}
open-ended evolution and supplies a principled bridge to the
physics of complexity that black-box drivers do not. Three
natural extensions, deferred to follow-up work, complete a
research programme that this paper opens but does not exhaust:
(i) a system-identified scale-separation timescale $\tau^*$,
(ii) a quantitative comparison between MSPD loss and
neural-network OEE losses on a shared task, and
(iii) multi-objective evolution combining MSPD with a
complementary snapshot-structural metric, which may dominate
either driver alone.

\begin{center}\rule{0.5\linewidth}{0.5pt}\end{center}

\subsection*{Data availability}\label{data-availability}

The data supporting the findings of this study were generated by
numerical simulation. The source code used to generate and analyse all
data, together with a selection of representative simulation snapshots
(checkpoints) underlying the reported claims, is available in a public
GitHub repository at \url{https://github.com/<org>/<repo>}. No
previously published or third-party datasets were used in this study.

\begin{center}\rule{0.5\linewidth}{0.5pt}\end{center}

% Keep [chan2019lenia] in the rendered bibliography even though it currently
% only appears in TODO comments (§2.2). Remove this \nocite once the
% Background section actually cites Chan in body text.
\nocite{chan2019lenia}

\bibliographystyle{abbrv}
\bibliography{references}

\subsubsection*{Appendix A. Proofs of the consistency statements}
\label{method:appendix-proof}

\paragraph{Proof of Proposition 1.}
All objects used in the proof are restated here.  Fix a finite set of windows \(k=1,\ldots,K\).  For each window, let \(m_k\) be a probability measure on local carriers, let \(L_{i,k}\) be the population transition law of carrier \(i\), and define
\[
  H_k=
  \iint D(L_{i,k},L_{j,k})\,m_k(di)m_k(dj).
\]
Let \(i_1,\ldots,i_n\) be sampled carriers and let \(\widehat L_{i_a,k}\) be empirical transition laws.  Define
\[
  \widehat H_k=
  \frac{2}{n(n-1)}
  \sum_{a<b}D(\widehat L_{i_a,k},\widehat L_{i_b,k}).
\]
By the pairwise plug-in consistency assumption,
\[
  \widehat H_k
  -
  \frac{2}{n(n-1)}
  \sum_{a<b}D(L_{i_a,k},L_{i_b,k})
  \xrightarrow[]{p}0.
\]
By the carrier sampling law of large numbers,
\[
  \frac{2}{n(n-1)}
  \sum_{a<b}D(L_{i_a,k},L_{i_b,k})
  \xrightarrow[]{p}
  H_k.
\]
Thus \(\widehat H_k\xrightarrow[]{p}H_k\).  The pooled-null assumption gives \(\widehat H_k^0\xrightarrow[]{p}0\), hence
\[
  \widehat{\Delta H}_k=\widehat H_k-\widehat H_k^0
  \xrightarrow[]{p}H_k.
\]
Since \(\phi\) is continuous at \(H_k\),
\[
  \phi(\widehat{\Delta H}_k)
  \xrightarrow[]{p}
  \phi(H_k).
\]
The number of windows is finite, so the empirical trace vector converges in probability to the population windowed trace vector.

Let \(a\in\mathbb R^K\).  The finite-grid functional has the form
\[
  \mathfrak M(a)=
  \frac{
    \sum_{j=1}^{J-1}w_j
    \frac{\|C_j a-U_jC_{j+1}a\|_2^2}
         {\|C_ja\|_2^2+\eta^2}
  }{
    \sum_{j=1}^{J-1}w_j
  },
\]
where \(C_j\) and \(U_j\) are fixed linear maps, \(w_j>0\), and \(\eta>0\).  This map is continuous on \(\mathbb R^K\), because every denominator is bounded below by \(\eta^2\).  The continuous mapping theorem gives convergence of the empirical MSPD to the windowed population target.

\paragraph{Proof of Proposition 2.}
Let \(h\in L^2([0,T])\).  For a window grid, let \(P_Kh\) denote the piecewise-window approximation whose value on each window is the average of \(h\) on that window.  For regular window grids with maximal diameter tending to zero,
\[
  P_Kh\to h
  \quad\text{in }L^2([0,T]).
\]
This is the standard convergence of local averages to the underlying \(L^2\) function.  The local-stationarity assumption in the proposition states that the population windowed trace differs from these local averages by a quantity tending uniformly to zero.  Therefore the windowed trace converges to \(h\) in \(L^2\).

The continuous pathwise MSPD is obtained by applying temporal coarse-graining, differentiating with respect to logarithmic scale, taking squared \(L^2\) norms, adding a positive denominator floor, and integrating over temporal scales.  The finite-grid functional applies discrete coarse-graining, a finite-difference approximation to the logarithmic scale derivative, and a finite quadrature over scales.  By the assumed convergence of the discrete coarse-graining operators and the scale quadrature to their continuous counterparts, and by the positive denominator floor, the finite-grid MSPD of the windowed trace converges to the continuous pathwise MSPD of \(h\).  This proves the deterministic approximation claim.

\subsection*{Appendix B. Experimental protocols and reproducibility details}
\label{appendix:protocols}

This appendix gives the complete execution protocol for the empirical claims in Section~4.  The main text gives the claim logic and the reported interpretation; this appendix gives the parameterization, sampling rules, rollout horizons, evaluation windows and statistical units needed to reproduce the measurements.

\subsubsection*{B.1 Claim-level pipeline}

\begin{table}[H]
\centering
\scriptsize
\begin{tabular}{p{0.08\linewidth}p{0.15\linewidth}p{0.27\linewidth}p{0.23\linewidth}p{0.17\linewidth}}
\toprule
Claim & Substrate & Pipeline & Reported output & Statistical unit \\
\midrule
N0 & synthetic particles & Generate known trajectory families; compute $\Delta H(w,\tau)$; aggregate MSPD; evaluate event and role diagnostics where ground truth exists & Family scores, heatmaps, event error, role recovery & synthetic family/seed \\
C1 & CA & Sweep all 18-bit rules; rank by transition-law MSPD; compare selected high-MSPD rules with random rules & optimized-vs-random transition-law MSPD & rule \\
C1 & Flow-Lenia & Optimize rule coordinates on fixed pathwise contexts; replay selected and matched random rules exactly; report split-B MSPD at the optimization-selected group lag & matched contrast $\Delta^{C1}_{s,r}$ & independent optimization run \\
C1 & Particle Life++ & Optimize neural interaction weights; sample matched random controls; run reusable lagrangian rollouts; evaluate fixed post-hoc lag on held-out windows & matched contrast $\Delta^{C1}_{s,r}$ & independent matched group \\
C2 & CA & Select branch boards from saved $\Delta H$ traces; perturb board sites; resume futures; measure Hamming divergence & association between $E_b$ and $B_b$ & branch state \\
C2 & Flow-Lenia & Select high/middle/low branch states from exact optimized trajectories; preserve the state; sample independent stochastic continuations; compare high- and low-energy future divergence & high-minus-low run contrasts aggregated across ranks and horizons & independent optimization/source run \\
C5 & Flow-Lenia & Branch exact visited states into matched free and temporary-wall continuations; preserve block/channel mass during confinement; subtract free/free divergence & excess-divergence contrast $\Delta^{C5}_{s,r}$ & independent matched optimization run \\
C5 & Particle Life++ & For each checkpoint, run $A_\theta$, $C_\theta$, and warmup-plus-cell-shuffle $W_\theta$; compare late futures & anchored frustration contrast $\Delta^{C5}_{s,r}$ & independent matched group \\
\bottomrule
\end{tabular}
\caption{End-to-end experimental pipeline for the reported claims.  Each row can be read independently: it states the source data, the intervention or optimization step, the reported statistic and the unit used for statistical aggregation.}
\label{tab:appendix-pipeline}
\end{table}
\FloatBarrier

\subsubsection*{B.2 Synthetic calibration protocol}
\label{app:synthetic-calibration-protocol}

The synthetic calibration uses particles on the unit torus with shortest-periodic displacement.  The local empirical law is the distribution of particle displacements inside a metric window and lag.  The default calibration uses $64$ particles, $240$ time steps, one seed per family, and lag grid
\[
  \mathcal T=\{1,2,4,8,12,16\}.
\]
MSPD is computed from processed $\Delta H(w,\tau)$ maps.  Families S0, S1 and S3 are low-score controls; S4 separates stationary role heterogeneity from temporal MSPD; S5, S6 and S8 test event localization; S7 and S8 test scale response.

\paragraph{Per-family analysis.}
S0 is the zero-motion null: all displacement laws collapse to the same point mass, so both processed $\Delta H$ and MSPD are zero up to numerical precision.  S1 contains stochastic motion without organized role structure; finite-window fluctuations produce a small processed $\Delta H$ but no structured temporal organization, giving a low MSPD score.  S3 is a coherent moving blob: visually organized but with nearly identical local transition laws, so both processed $\Delta H$ and MSPD remain near zero.  This is the negative control showing that coherent object motion is not by itself complexity for MSPD.

S4 plants stationary role heterogeneity: different particles have persistently different transition laws, so processed $\Delta H$ amplitude is high, but MSPD is much smaller because the role structure is essentially stationary across windows; this separates mean heterogeneity from temporally organized complexity.  S5 plants a synchronous global switch: the change is shared by nearly all particles at once, so windowed local transition laws remain nearly homogeneous and the MSPD response is low.  This is a resolution-dependent failure case: MSPD is only sensitive to temporal structure resolved by the chosen window and lag grid.  Reducing window length would help but would also reduce displacement samples per $\Delta H_{k,\tau}$; the appropriate remedy is to increase the logged temporal resolution so shorter physical windows can be used while retaining enough samples.

S6 uses staggered transition times: the transition is spread across particles and windows so local transition laws differ in a temporally extended way, producing both visible $\Delta H$ event structure and a high MSPD score.  S7 contains multi-scale moving groups: it gives the largest processed $\Delta H$ amplitude because persistent groups have clearly different transition laws, but its MSPD is lower than S8 because much of this organization is stable rather than reorganizing over time.  S8 combines a split event with role emergence: the system starts from a coherent object and separates into multiple groups with different motion regimes, producing the largest MSPD score even though its mean $\Delta H$ amplitude is below S7.  This case illustrates the intended distinction: MSPD rewards temporally organized changes in local heterogeneity, not simply the largest average heterogeneity.

\subsubsection*{B.3 Random-control sampling}

Random controls are sampled separately for each substrate and matched to the same reporting group as the optimized checkpoint.  The sampling rule is part of the null model: it defines what counts as an unoptimized parameter from the same substrate family.

\begin{table}[H]
\centering
\footnotesize
\begin{tabular}{p{0.18\linewidth}p{0.32\linewidth}p{0.39\linewidth}}
\toprule
Substrate & Random parameter $\theta^{\mathrm{rand}}$ & Matching rule \\
\midrule
Life-like CA & Uniform random 18-bit totalistic birth/survival rule & Compared with selected high-MSPD rules under the same board and scoring protocol \\
Flow-Lenia & Three independent rule-offset vectors sampled from the optimizer's zero-centered initialization law and scale & Each random rule is paired with the optimized run and evaluated at the same group lag; C1 also matches the realised initial physical state and tracer initialization \\
Particle Life++ & Fresh neural interaction-weight vector from the substrate-default initialization distribution & Three random checkpoints are paired with each optimized group in C1 and C5 \\
\bottomrule
\end{tabular}
\caption{Random-control sampling.  Random controls are not resampled after seeing the post-hoc score; they are evaluated by the same reported protocol as the optimized checkpoints.}
\label{tab:random-controls}
\end{table}
\FloatBarrier

\subsubsection*{B.4 Optimization protocols}

\begin{table}[H]
\centering
\scriptsize
\begin{tabular}{p{0.14\linewidth}p{0.27\linewidth}p{0.24\linewidth}p{0.25\linewidth}}
\toprule
Substrate & Optimized object and search & Rollout and metric budget during optimization & Optimization objective \\
\midrule
CA & Exhaustive enumeration of $2^{18}$ totalistic rules & transition-law MSPD on simulated boards; active-site weighting & rank rules by transition-law MSPD and evaluate selected rules on holdout boards \\
Flow-Lenia & Mirrored OpenAI-ES over rule offsets plus one lag-selector coordinate; eight candidates per iteration; $100$ iterations; search scale $0.2$ & grid $128$; four fixed pathwise contexts per run; rollout length $300{,}000$; metric range $50{,}000$--$300{,}000$; window $20{,}000$; step $5{,}000$; lag grid $\{1000,2000,\ldots,10000\}$; 48 displacement samples; 16 projections; six null repetitions; 64 sampled tracers & maximize mean MSPD over the four fixed contexts; amplitude term disabled \\
Particle Life++ & Sep-CMA-ES over neural pair-interaction weights; population size $8$; $500$ iterations; initial scale $\sigma=0.2$ & $600$ particles; six colors; rollout length $6000$; window $100$; step $50$; range $0$--$1000$; lag grid $\{5,10,15,20,25,30,35\}$; 48 displacement samples; 16 projections; six null repetitions; 64 particle samples & minimize $-(\lambda_M\operatorname{MSPD}+\lambda_H\overline{\Delta H})$ with nonzero mean-heterogeneity weight to avoid stationary random-weight basins \\
\bottomrule
\end{tabular}
\caption{Optimization protocols.  These are optimization-time estimators; the reported C1 scores are computed by the post-hoc protocols in Appendix~B.5.}
\label{tab:optimization-protocols}
\end{table}
\FloatBarrier

Particle Life++ optimization uses $dt=0.02$, half-life $0.04$, interaction radius $r_{\max}=0.1$, repulsion boundary $\beta=0.3$, particle mass $0.1$, render radius $0.04$, sharpness $30.0$, toroidal boundary conditions and dense neighbor interactions.  Color updates are enabled.

\subsubsection*{B.5 C1 post-hoc evaluation protocols}

\begin{table}[H]
\centering
\scriptsize
\begin{tabular}{p{0.15\linewidth}p{0.23\linewidth}p{0.24\linewidth}p{0.28\linewidth}}
\toprule
Substrate & Rollout source & Metric grid and windowing & Reported score \\
\midrule
CA & holdout board simulations for selected and random rules & categorical transition laws; active-site weighting; same board/evaluation protocol for selected and random rules & transition-law MSPD \\
Flow-Lenia & exact optimizer-native replay of four fixed trajectories for each selected rule and matched random controls & grid $128$; horizon $300{,}000$; range $50{,}000$--$300{,}000$; window $20{,}000$; step $5{,}000$; lag grid $\{1000,2000,\ldots,10000\}$; 8192 logged tracers, 64 sampled per metric cell & fixed optimization-selected group lag; split A is diagnostic and split B is the reporting partition \\
Particle Life++ & reusable single-trajectory lagrangian rollouts with total horizon $24{,}000$ & evaluate steps $4000$--$24{,}000$; fixed $\tau=25$; window $100$; step $50$; 48 samples; 16 projections; six null repetitions; 64 particles & held-out fixed-lag MSPD on odd-indexed metric windows \\
\bottomrule
\end{tabular}
\caption{C1 post-hoc evaluation.  The same post-hoc score is applied to optimized checkpoints and matched random controls within a substrate.}
\label{tab:c1-posthoc}
\end{table}
\FloatBarrier

The interleaved temporal split remains useful because these systems are nonstationary: early and late windows can represent different regimes.  For Particle Life++ it retains its existing selection/evaluation role.  For Flow-Lenia, however, the lag is inherited from the selected search vector and fixed for every member of the matched group.  Even-indexed split A is diagnostic and odd-indexed split B is the reporting partition; split B is not a fresh-seed or fresh-initial-state holdout from the optimization objective.

For each Flow-Lenia run, the selected optimized rule is replayed on its four original evaluation keys.  Three matched random rules are inserted into unused optimizer-native population lanes so that batching and key-splitting topology match the optimized replay.  For each rollout context, the random rule retains its own rule-dependent kernel and growth fields, while its returned initial $A/P$ state is overwritten with the exact optimized candidate's returned initial state; tracer positions and channels are then initialized with the same keys.  At the fixed group lag, the optimized group value is the median of four split-B rollout scores and the random group value is the median of all twelve random-rule-by-context scores.  The ten run-level contrasts are the independent inferential units.

\paragraph{Substrate-specific C1 results.}
\label{app:c1-substrate-results}
\emph{Life-like CA.}  The CA experiment evaluates the selected high-MSPD rules against random rules using the same categorical transition-law MSPD score.  The selected rules are separated from random controls, confirming that the estimator can be optimized in a substrate without a velocity variable.  This result should be interpreted relative to the evaluation distribution: MSPD is measured over rollouts generated from the specified initialization protocol, not over all possible configurations of a rule.  Consequently, even computationally universal systems such as Life need not have high MSPD under a given initialization ensemble; the score reflects the complexity expressed by typical sampled rollouts, rather than the maximal behavior that the rule can realize under carefully constructed initial conditions.

\emph{Flow-Lenia.}  For Flow-Lenia, C1 asks whether search found rule parameters with larger MSPD on the exact fixed pathwise objectives used during optimization than independent matched random rules from the same initialization distribution.  The optimized and random rules are replayed from the same realised physical state and tracer initialization within each rollout context, and all members of a matched group use the optimized candidate's lag.  The split-B group contrast is positive in $9/10$ independent runs, with median $4.69\times10^{-4}$ and exact one-sided sign-test $p=0.0107$.  This supports successful optimization under the fixed-objective protocol; it does not establish generalization to unseen initial states or unseen stochastic realizations.

\emph{Particle Life++.}  Particle Life++ gives a statistically confirmed optimized-vs-random MSPD increase.  This large contrast is also mechanistically interpretable.  Many random neural interaction networks enter low-motion stationary regimes after damping, which makes particle displacement laws narrow and similar across windows.  Optimized checkpoints leave this basin and maintain moving particle groups with different interaction histories.  The measured increase is therefore both a post-hoc MSPD increase and a contrast against a random-weight baseline with a strong stationary attractor.

\emph{Lag and temporal-split audit.}  Particle Life++ retains the post-hoc lag-selection procedure described above.  Flow-Lenia instead inherits one lag from the optimized search vector and applies it to the optimized candidate and all three random controls in the group.  Split A and split B are interleaved diagnostic and reporting partitions of the same exact replay.  This preserves a temporally separated reporting score without treating split B as fresh-seed or fresh-initial-state generalization.

\begin{figure}[H]
\centering
\begin{minipage}{0.48\linewidth}
  \centering
  \includegraphics[width=\linewidth]{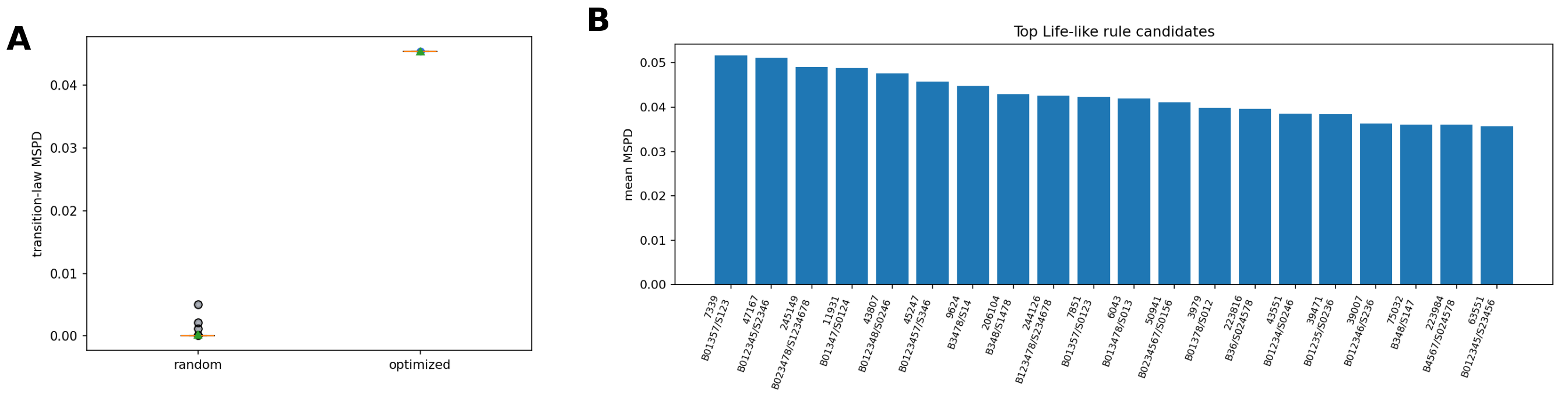}
  \subcaption{Cellular-automata C1.  Left: post-hoc MSPD for matched random rules and the optimized/selected rule.  Right: MSPD scores of the top Life-like rule candidates from the exhaustive rule sweep.}
  \label{fig:ca-c1}
\end{minipage}\hfill
\begin{minipage}{0.48\linewidth}
  \centering
  \includegraphics[width=\linewidth]{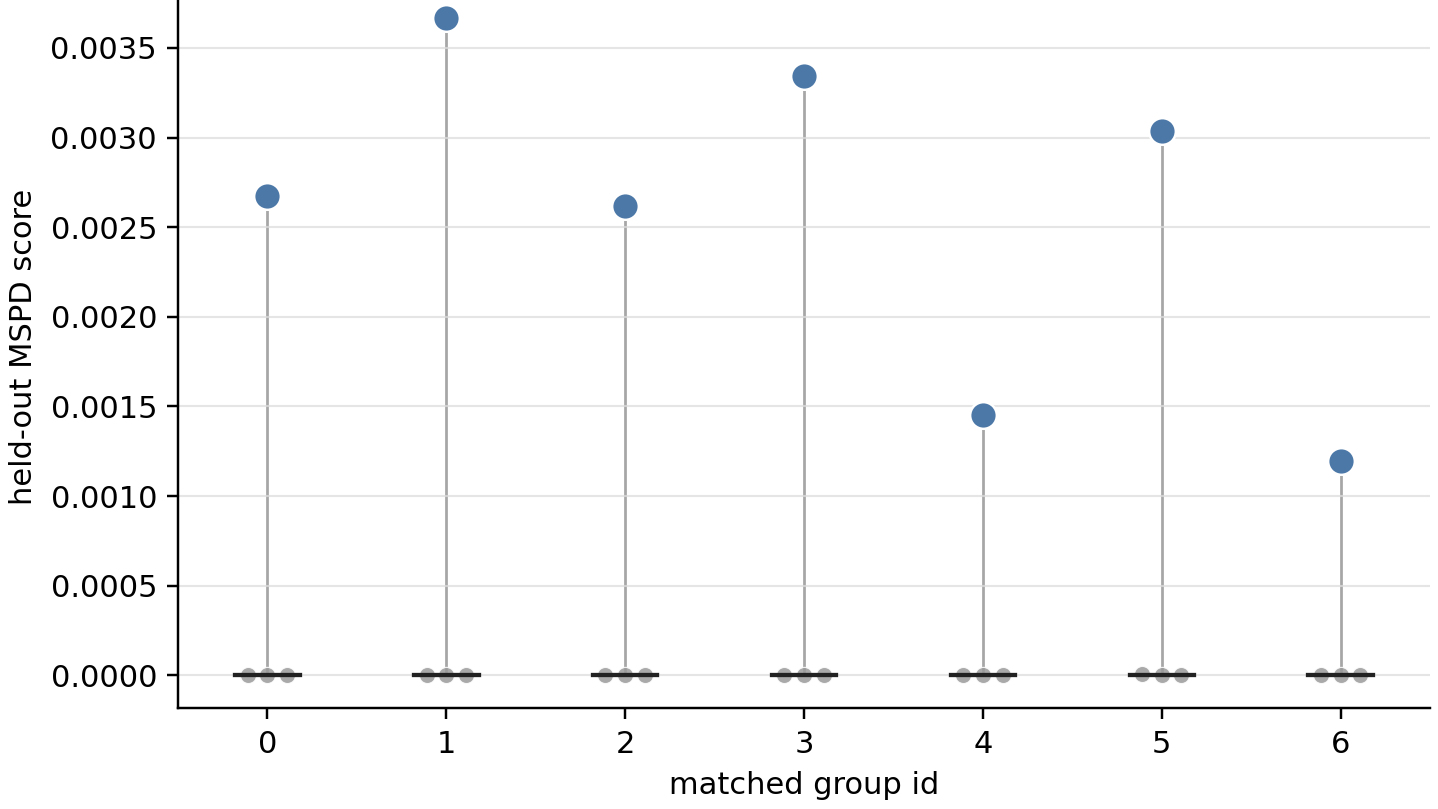}
  \subcaption{Particle Life++ C1.  Blue points are optimized checkpoints, gray points are matched random controls, and black marks denote random-control medians.  All matched groups are positive.}
  \label{fig:c1-plife}
\end{minipage}
\caption{C1 results for CA (a) and Particle Life++ (b).}
\end{figure}
\FloatBarrier

\begin{figure}[H]
\centering
\includegraphics[width=0.96\linewidth]{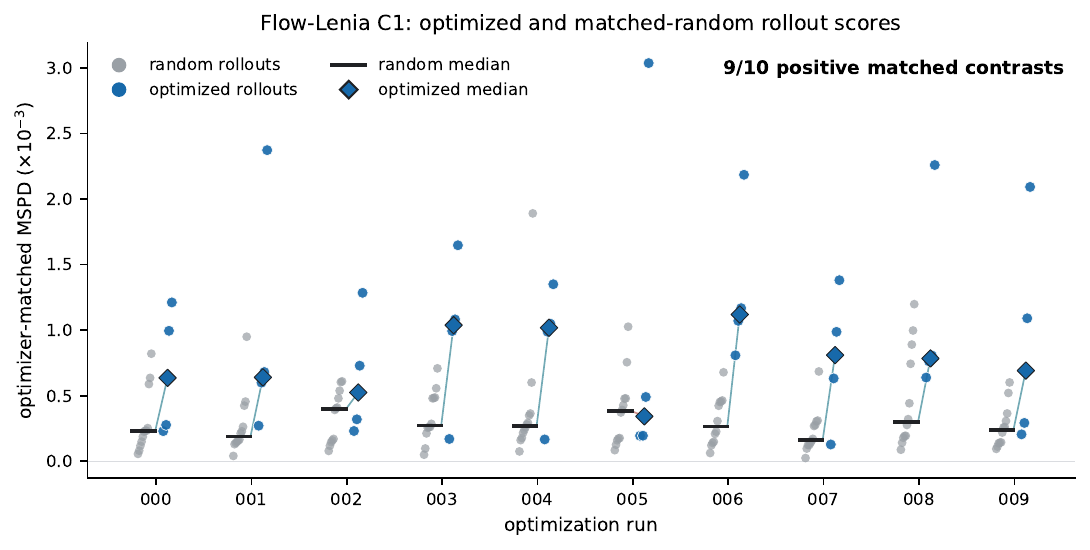}
\caption{Flow-Lenia C1 under optimizer-matched exact replay.  Each group is one optimization run.  Blue points are the four optimized-candidate rollout scores, gray points are the twelve matched random-control rollout scores, the diamond is the optimized median, and the black segment is the random-control median.  The optimized-minus-random-median contrast is positive in $9/10$ groups.  This is an exact fixed-context replay comparison, not a fresh-initial-state or fresh-seed generalization test.}
\label{fig:c1-flow}
\end{figure}
\FloatBarrier

\begin{figure}[H]
\centering
\includegraphics[width=0.96\linewidth]{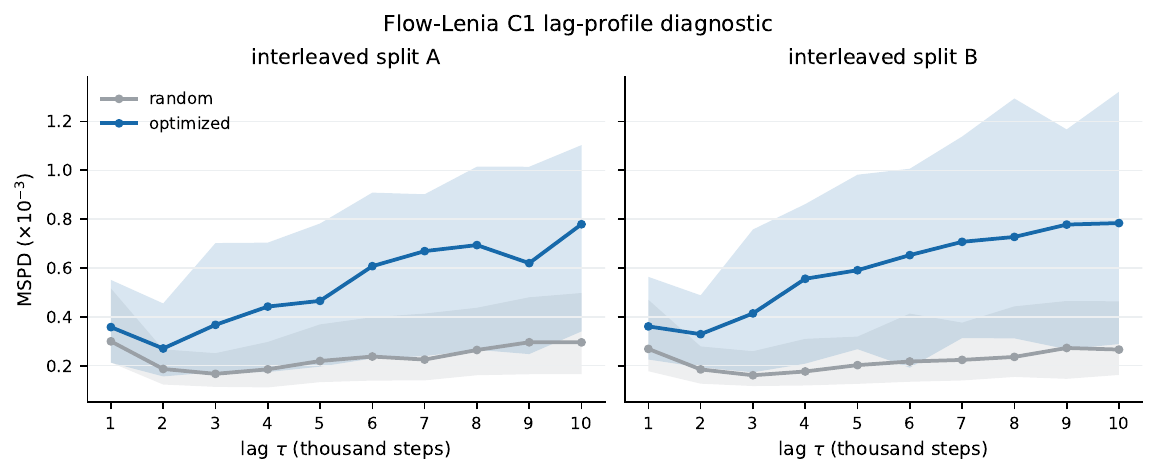}
\caption{Flow-Lenia C1 lag-profile diagnostic.  Curves show median MSPD and shaded interquartile ranges across rollout-level scores for optimized and matched random candidates.  Panels A and B are the two interleaved temporal partitions, split A and split B.  They are diagnostic partitions rather than lag-selection and held-out-evaluation panels; the reported group comparison inherits the lag of the optimization-selected candidate.}
\label{fig:c1-tau-profiles}
\end{figure}
\FloatBarrier

\subsubsection*{B.6 C2 branch-continuation protocols}
\label{app:c2-branch-perturbations}

For every branch state $b$, the C2 pipeline computes branch energy $E_b$, generates controlled alternative futures and computes future divergence $B_b$.  The continuation mechanism is substrate-specific, but the rule parameters are fixed in both substrates.

\begin{table}[H]
\centering
\scriptsize
\begin{tabular}{p{0.15\linewidth}p{0.25\linewidth}p{0.27\linewidth}p{0.23\linewidth}}
\toprule
Substrate & Branch sampling & Continuation family & Outcome distance \\
\midrule
CA & 24 branch windows sampled across the saved $\Delta H$ distribution & four continuations per branch; independently flip fraction $0.01$ of lattice sites; resume the same rule for 64 steps & median pairwise future Hamming divergence \\
Flow-Lenia & ten exact optimized source trajectories; five high, five middle and five low states per run & three exact-state continuations per branch; no state noise; use an independent continuation seed while preserving the exact state & median pairwise CLIP--Chamfer distance between seven-frame future clouds \\
\bottomrule
\end{tabular}
\caption{C2 branch-continuation protocol.  CA creates alternative futures by perturbing a deterministic state.  Flow-Lenia preserves the exact state and samples the simulator's intrinsic stochastic continuation law.}
\label{tab:c2-protocols}
\end{table}
\FloatBarrier

For Flow-Lenia, C2 uses the selected optimized candidate and rollout-context index zero from each of the ten optimization runs.  The full $\Delta H$ map is computed at all ten lags.  For each eligible window, branch energy is
\[
  E_b=\frac{1}{10}\sum_{\tau\in\{1000,\ldots,10000\}}\max(\Delta H(w_b,\tau),0).
\]
Eligible window centers run from $60{,}000$ through $280{,}000$ in $5{,}000$-step increments.  Within each source trajectory, the high pool is the top $20\%$ of branch energies, the low pool is the bottom $20\%$, and the middle pool spans empirical quantile ranks $40\%$--$60\%$.  Five states are sampled without replacement from each stratum with at least $5{,}000$ steps between selected branch times, giving $15$ states per run and $150$ source states in total.  For source-trajectory order $o$, the selection seed is $12345+10007o$.  For selected point index $p$, branch index $m$ and stratum $c$, the continuation seed is
\[
  12345+1000003o+1009p+131m+\delta_c,
\]
with $\delta_{\rm high}=0$, $\delta_{\rm mid}=3967$ and $\delta_{\rm low}=7919$.

At a selected branch time, the exact saved optimizer-native state and execution context are loaded.  The selected rule, activity field, internal field, tracers and population lane are preserved.  External perturbation scales are exactly zero,
\[
  A_{\rm std}=P_{\rm std}=q_{\rm std}=0,
\]
and only the continuation seed is changed.  The stored branch-start audit is bit exact for $A$ and $P$ in all $450$ branches.

Each branch is simulated once to $30{,}000$ steps.  The reported horizons are
\[
  T\in\{5{,}000,10{,}000,15{,}000,20{,}000,30{,}000\},
\]
with eight retained frames per horizon.  The retained relative steps are
\begin{center}
\footnotesize
\begin{tabular}{c l}
\toprule
Horizon & Retained relative steps \\
\midrule
$5$k & $0,700,1400,2100,2850,3550,4250,5000$ \\
$10$k & $0,1400,2850,4250,5700,7100,8550,10000$ \\
$15$k & $0,2100,4250,6400,8550,10700,12850,15000$ \\
$20$k & $0,2850,5700,8550,11400,14250,17100,20000$ \\
$30$k & $0,4250,8550,12850,17100,21400,25700,30000$ \\
\bottomrule
\end{tabular}
\end{center}
The common offset-zero frame is excluded, leaving seven-frame embedding clouds.  For state $b$ and horizon $T$, $B_b(T)$ is the median of the three pairwise CLIP--Chamfer distances among the three branch futures.

Within each run, high and low states are paired only by deterministic local draw order.  For rank $p$ and horizon $T$,
\[
  D_{r,p,T}=B_{\rm high}(r,p,T)-B_{\rm low}(r,p,T).
\]
The run statistic averages the five ranks and five horizons.  The independent unit is the optimization run, not the branch state or horizon.  All ten run aggregates are positive; the exact one-sided sign and signed-rank tests are therefore both $p=0.0009766$.  Horizon-wise summaries are reported as sensitivity diagnostics rather than independent replications.

\begin{figure}[H]
\centering
\includegraphics[width=\textwidth,height=0.72\textheight,keepaspectratio]{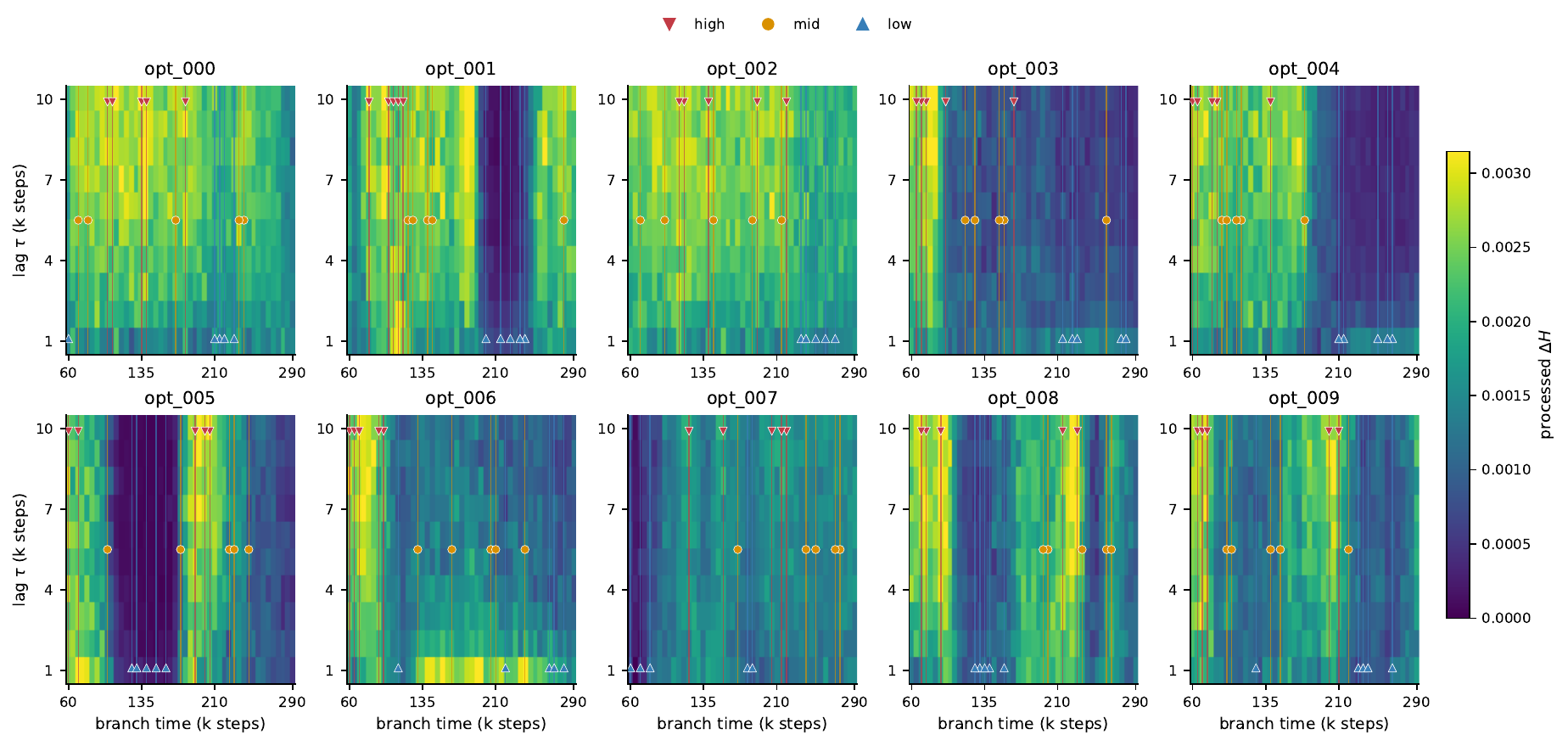}
\caption{Flow-Lenia C2 branch-selection preview for all ten source optimization runs.  Each panel shows the processed $\Delta H$ heatmap and the five selected high, five middle and five low branch times.  Marker height is a fixed display lane for the stratum and is not a selected lag $\tau$; branch times are selected from branch energy aggregated over the admissible lag set.  The strata define the matched high-minus-low sampling design.}
\label{fig:c2-branch-selection-preview}
\end{figure}

Figure~\ref{fig:c2-branch-selection-preview} documents the stratified branch-state selection used by the current C2 analysis.
\FloatBarrier

\paragraph{Substrate-specific C2 results.}
\label{app:c2-substrate-results}
\emph{Cellular automata.}  For CA, $E_b$ is computed from categorical transition-law $\Delta H$.  A branch state is the saved binary board $x_t$ at the selected branch time.  For each continuation $m=1,\ldots,M$, we draw a subset $S_m\subset\Lambda$ of lattice sites uniformly without replacement and flip those sites before resuming the same Life-like rule:
\[
  x^{(m)}_t(u)=
  \begin{cases}
    1-x_t(u), & u\in S_m,\\
    x_t(u), & u\notin S_m .
  \end{cases}
\]
The subset size is fixed as a small fraction of the board; the numerical value used in the reported experiment is given above.  Each perturbed board is then evolved forward with the same rule and the same boundary convention.  The future distance between two continuations is normalized Hamming distance between the resumed boards, aggregated over the continuation horizon,
\[
  d_{\mathrm{CA}}(Y_p,Y_q)
  =
  \frac{1}{H}
  \sum_{h=1}^{H}
  \frac{1}{|\Lambda|}
  \sum_{u\in\Lambda}
  \mathbf 1\!\left[
    x^{(p)}_{t+h}(u)\ne x^{(q)}_{t+h}(u)
  \right],
\]
and the branch sensitivity score is $B_b=\operatorname{median}_{p<q}d_{\mathrm{CA}}(Y_p,Y_q)$.

\emph{Flow-Lenia.}  Flow-Lenia provides the quantitative C2 test.  Each branch starts from an exact saved optimizer-native state.  The activity field, internal field and tracer state are copied without external perturbation; only the continuation seed is changed, so the branches sample $Y_{b,m}\sim P(\text{future}\mid X_b,\phi)$ under the simulator's intrinsic stochastic continuation law.  No Gaussian noise is added to $A$, $P$ or tracer coordinates.  Within each of the ten optimized source trajectories, branch states are sampled from low, middle and high within-trajectory branch-energy strata; the primary comparison uses five high and five low states, with three stochastic continuations per state.  State divergence is the median pairwise CLIP--Chamfer distance between future embedding clouds.

\begin{figure}[H]
\centering
\begin{subfigure}[t]{0.32\linewidth}
  \centering
  \includegraphics[width=\linewidth]{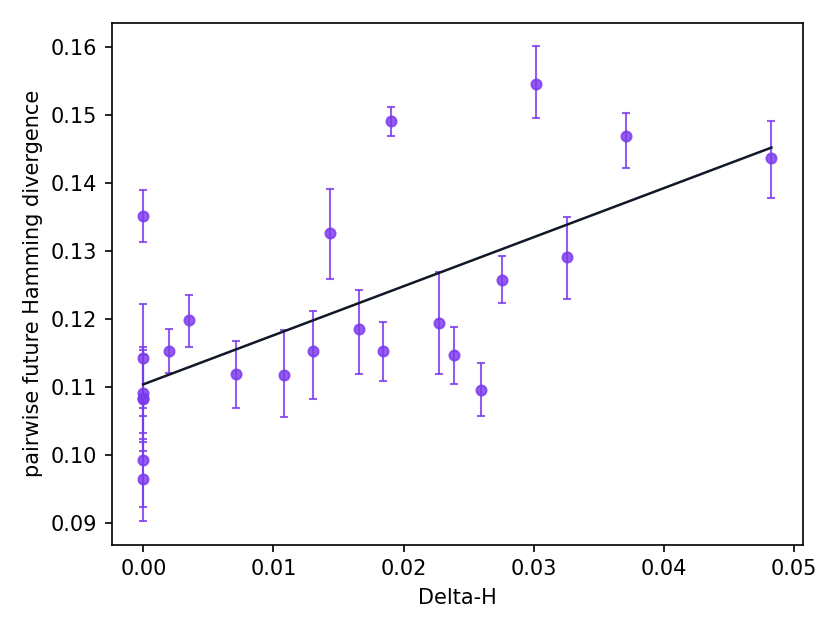}
  \subcaption{CA C2.  Each point is a sampled branch state; horizontal axis is transition-law $\Delta H$ at the branch window; vertical axis is median pairwise Hamming divergence between perturbed continuations.}
  \label{fig:ca-c2}
\end{subfigure}\hfill
\begin{subfigure}[t]{0.66\linewidth}
  \centering
  \includegraphics[width=\linewidth]{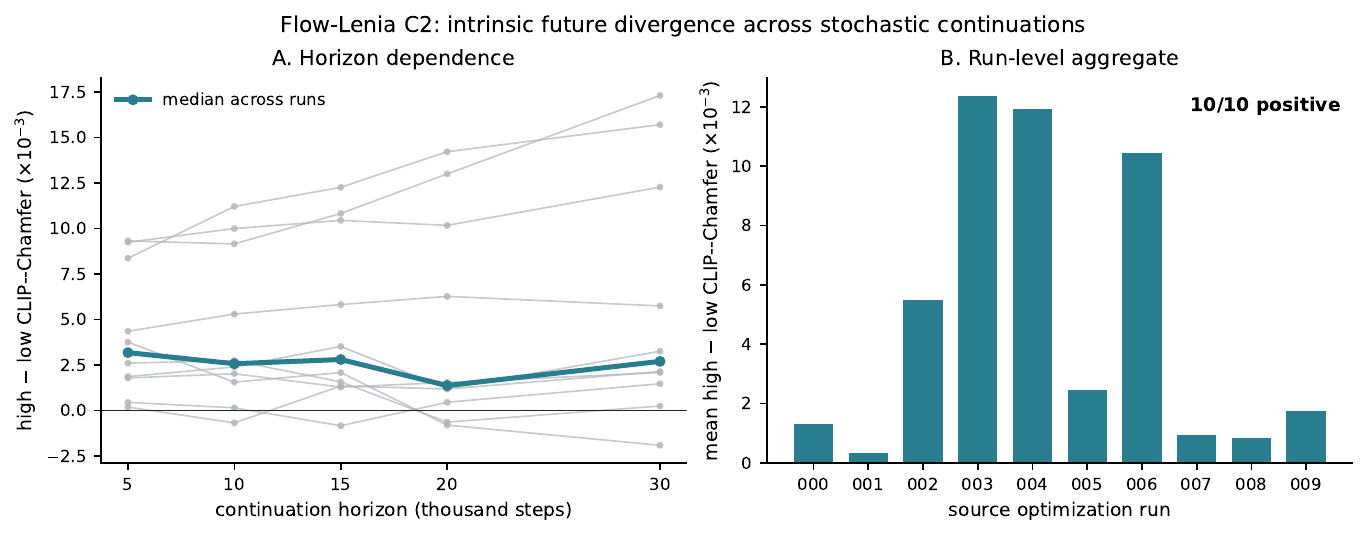}
  \subcaption{Flow-Lenia C2 under exact-state stochastic continuations.  Left panel: run-level mean matched high-minus-low future CLIP--Chamfer divergence at horizons $5$k--$30$k steps; thin lines are the ten source runs, thick line is their median.  Right panel: each run's aggregate over five matched branch ranks and five horizons.  All $10/10$ run aggregates are positive.}
  \label{fig:c2-flow}
\end{subfigure}
\caption{C2 results for CA (a) and Flow-Lenia (b).  The shared branch-point frame is excluded from the Flow-Lenia distance.}
\end{figure}

\emph{Interpretation and boundary.}  The CA experiment is a discrete-substrate check of the same C2 logic.  The positive trend in Fig.~\ref{fig:ca-c2} indicates that branch states with larger categorical transition-law heterogeneity tend to produce larger divergence after small changes to the current board; the C2 readout is not specific to continuous Flow-Lenia fields or to a neural image-distance metric.  For Flow-Lenia, exact-state stochastic continuations from high-$\Delta H$ states diverge more than continuations from low-$\Delta H$ states in all ten independent source runs after aggregation over the reported horizons.  The result identifies $\Delta H$ as a state-level marker of intrinsic future divergence along the sampled optimized trajectories; it does not establish a universal uncertainty measure for every Flow-Lenia rule or state distribution.

\FloatBarrier

\subsubsection*{B.7 C5 blockwise-frustration protocols}

C5 interrupts early spatial co-organization while holding the checkpoint fixed, releases the system, and compares post-release futures after subtracting ordinary future variability.  The Particle Life++ protocol is unchanged.  The final Flow-Lenia protocol is as follows.

Flow-Lenia uses the same ten optimized and thirty matched random rule candidates as C1.  Each candidate branches from its own exact C1 rollout-context-zero trajectory.  The fifteen absolute branch times and three branch seeds selected from the optimized C2 source run are reused for the optimized candidate and all three matched random candidates in the group.  Thus candidate-specific state selection cannot create the optimized-versus-random contrast.  The complete plan contains $40$ candidates, $15$ points and $3$ branch seeds, for $1800$ free plans and $1800$ wall plans.

For candidate $c$, branch point $b$, branch seed $m$ and horizon $T$, the free and wall arms start from the same saved state and use the same continuation seed.  No external state noise is added.  The domain is partitioned into a $3\times3$ array of isolated compartments with five-cell hard-zero padding.  Walls are active for the first half of the horizon and then released.  The release steps for horizons $5$k, $10$k, $15$k, $20$k and $30$k are respectively $2500$, $5000$, $7500$, $10000$ and $15000$.  One wall-constrained prefix is simulated to $15000$ steps and exact full-state snapshots at these release points are forked into the corresponding global continuations.  One global mutation event and one global categorical-reintegration field are generated per simulator step and partitioned over the blocks, so the nine compartments do not receive nine independent stochastic events.

During confinement, a block/channel mass projection is applied after every transition.  For each block and activity channel,
\[
  A_{\rm next}\leftarrow A_{\rm next}
  \frac{\text{mass before the step}}{\text{retained mass after RT and masking}},
\]
with guarded handling of empty channels.  This projection applies only to $A$ during the wall phase; it is not part of ordinary C1 or C2 dynamics.  Its purpose is to prevent the wall intervention from becoming a mass-extinction intervention.

For each horizon, the last four retained frames occur strictly after wall release.  The offsets are $2850,3550,4250,5000$ at $5$k; $5700,7100,8550,10000$ at $10$k; $8550,10700,12850,15000$ at $15$k; $11400,14250,17100,20000$ at $20$k; and $17100,21400,25700,30000$ at $30$k.  Let $F_m$ be the free future and $W_m$ the wall-then-release future for branch seed $m$.  The point-level intervention distance and intrinsic free spread are
\[
  I_b(T)=\operatorname{median}_m d_{\rm Ch}(F_m,W_m),
  \qquad
  V_b(T)=\operatorname{median}_{m<n}d_{\rm Ch}(F_m,F_n),
\]
and the excess divergence is
\[
  Q_b(T)=I_b(T)-V_b(T).
\]
For each candidate, $Q_{\rm candidate}(T)$ is the median over its fifteen branch points.  The run contrast is the optimized value minus the median of the three matched random candidates.

The $5{,}000$-step sensitivity horizon shows $8/10$ run contrasts positive, with median $5.83\times10^{-4}$, one-sided sign-test $p=0.0547$, and signed-rank $p=0.0967$.  This is directional evidence.  The frozen analysis designated $20{,}000$ steps as the confirmatory horizon; that test was $7/10$ positive and did not reject the null.  The horizon grid reuses prefixes of the same continuations and must not be interpreted as independent replication.

\paragraph{Substrate-specific C5 results.}
\label{app:c5-substrate-results}
\emph{Flow-Lenia.}  Flow-Lenia C5 branches from exact visited C1 states.  For each point and continuation seed, the free and wall arms start from the same saved state and use the same continuation seed.  During the first half of the continuation, a temporary $3\times3$ compartmentalization suppresses cross-region interactions; the block-and-channel mass projection above prevents the intervention from becoming a mass-extinction test.  After wall release, ordinary global dynamics resume.  The point score is the median same-seed free/wall CLIP--Chamfer distance minus the median CLIP--Chamfer divergence among free stochastic continuations.  At the local $5{,}000$-step sensitivity horizon, the optimized-minus-random contrast is positive in $8/10$ runs.

\emph{Particle Life++.}  Particle Life++ uses a cell-shuffle intervention rather than walls because particles are not arranged on a fixed spatial field.  After a free warmup, the plane is divided into cells and the cell arrangement is permuted while preserving within-cell coordinates and velocities.  This disrupts large-scale spatial organization without injecting a direct velocity kick.  The result is positive in $6/7$ matched groups with median contrast $0.00477$ and one-sided sign-test $p=0.0625$.

\begin{figure}[H]
\centering
\begin{subfigure}[t]{0.62\linewidth}
  \centering
  \includegraphics[width=\linewidth]{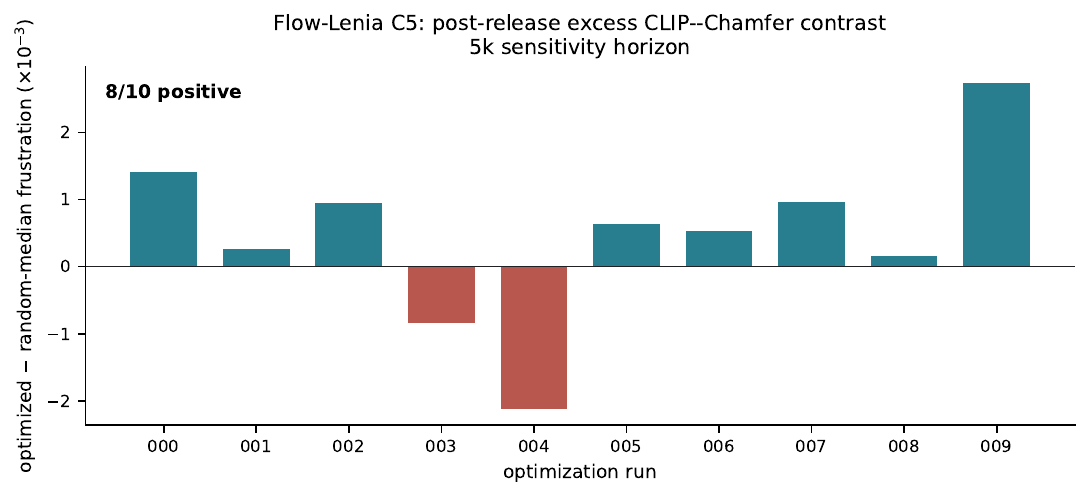}
  \subcaption{Flow-Lenia C5 at the $5$k sensitivity horizon.  Each bar is the optimized candidate's post-release excess CLIP--Chamfer score minus the median score of three matched random candidates.  Candidate values are medians over fifteen matched branch points.  Positive in $8/10$ runs.  The $5$k readout is a sensitivity analysis rather than the pre-designated confirmatory horizon.}
  \label{fig:c5-flow}
\end{subfigure}\hfill
\begin{subfigure}[t]{0.36\linewidth}
  \centering
  \includegraphics[width=\linewidth]{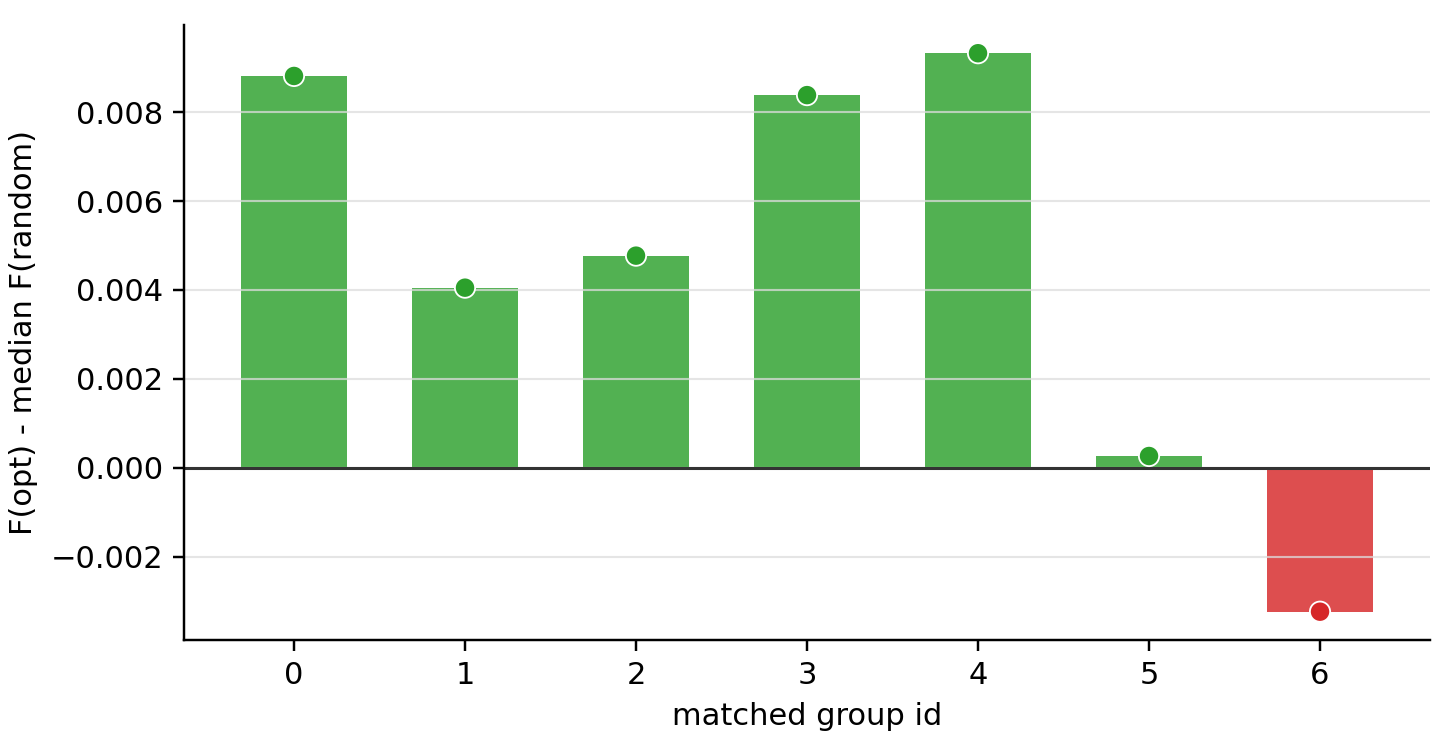}
  \subcaption{Particle Life++ C5.  Each bar is the optimized-minus-random matched contrast of the anchored frustration score $F_s(\theta)=d(A,W)-d(A,C)$.  Sign pattern is directionally consistent with stronger history dependence in optimized checkpoints.}
  \label{fig:c5-plife}
\end{subfigure}
\caption{C5 results for Flow-Lenia (a) and Particle Life++ (b).}
\end{figure}
\FloatBarrier

\subsubsection*{B.8 Reported statistics}

C1 and C5 use matched-group sign tests because the independent unit is the optimization group and the group counts are small.  For $n$ matched groups and $n_+$ positive contrasts, the one-sided sign-test value is
\[
  p=\Pr[\operatorname{Binomial}(n,1/2)\ge n_+].
\]
CA C2 reports a continuous branch-state association.  Flow-Lenia C2 reports directional tests on ten run-level high-minus-low aggregates; branch points, branch pairs and horizons inside a run are repeated measures.  C1 and C5 likewise use the independent optimization run as the statistical unit.

\subsubsection{B.9 Flow-Lenia simulator, update and RT tracer logging}
\label{app:flow-lenia-algorithm}

\paragraph{Parameter semantics.}
The optimizer-updated rule coordinates are denoted $\phi\in\mathbb R^{110}$.  They are raw-logit offsets around a deterministic base rule.  A scalar nuisance coordinate $u_\tau$ selects one of ten MSPD lags, so the search vector is $\vartheta=(\phi,u_\tau)\in\mathbb R^{111}$.  A realised checkpoint is $\theta=(\phi,\chi)$, where $\chi$ contains the returned initial state, random seeds and their exact propagation schedule, simulator configuration and execution context needed to reproduce the numerical path.  Only $\phi$ and $u_\tau$ are optimizer-updated.

\begin{table}[H]
\centering
\footnotesize
\begin{tabular}{p{0.42\linewidth}p{0.42\linewidth}}
\toprule
Setting & Canonical value \\
\midrule
Grid and boundary & $128\times128$, wall \\
Activity/internal channels & $3$ activity channels, $9$ internal fields \\
Routed kernels & $9$ kernels, $3$ radial components each \\
Routing matrix & $[[2,1,0],[0,2,1],[1,0,2]]$ \\
Integration & $d_d=5$, $dt=0.2$, effective $\sigma_{\rm RT}=0.2$ \\
Hidden-state mixing & categorical mode \\
Mutation & probability $0.05$ per step, $40\times40$ patch, scale $1.0$ \\
Disabled mechanisms & food, volcano, baseline mass decay, clipping, explicit renormalization \\
\bottomrule
\end{tabular}
\caption{Canonical Flow-Lenia simulator settings used by the ten optimization runs and exact replay experiments.}
\label{tab:flow-simulator-settings}
\end{table}
\FloatBarrier

\begin{table}[H]
\centering
\footnotesize
\begin{tabular}{p{0.36\linewidth}p{0.16\linewidth}p{0.28\linewidth}}
\toprule
Rule component & Count & Physical range \\
\midrule
Global radius $R$ & $1$ & $[2,25]$ \\
Relative radii $r$ & $9$ & $[0.2,1]$ \\
Growth centers $m$ & $9$ & $[0.05,0.5]$ \\
Growth widths $s$ & $9$ & $[0.001,0.18]$ \\
Radial centers $a$ & $27$ & $[0,1]$ \\
Radial amplitudes $b$ & $27$ & $[0.001,1]$ \\
Radial widths $w$ & $27$ & $[0.01,0.5]$ \\
Food-consumption coordinate & $1$ & $[0,0.5]$; inactive because food is disabled \\
\bottomrule
\end{tabular}
\caption{Decoded Flow-Lenia rule parameterization.  Raw offsets are added to a fixed base rule in logit coordinates and mapped through the sigmoid to these ranges.}
\label{tab:flow-rule-ranges}
\end{table}
\FloatBarrier

\paragraph{Simulator settings and initialization.}
The canonical grid is $128\times128$ with wall boundary, three nonnegative activity channels $A$, nine internal fields $P$, nine routed kernels and three radial components per kernel.  The integration timestep is $dt=0.2$, $d_d=5$, and the audited effective RT width is $\sigma_{\rm RT}=0.2$ in all ten runs.  Food and volcano mechanisms, baseline mass decay, explicit mass renormalization and mass clipping are disabled.  Hidden-state mixing uses categorical mode.  In ordinary global simulation without an externally supplied Gumbel tensor, the implementation uses a fixed categorical key rather than independently resampling this field at every step.  The visible pathwise stochasticity is supplied primarily by mutation: at each step, with probability $0.05$, one random $40\times40$ patch receives a shared nine-dimensional standard-normal increment.

Each rollout starts from four random $20\times20$ patches on the grid.  Patch overlap is allowed.  Activity values are sampled independently from $\mathrm{Unif}[0,1]$; one nine-dimensional internal-state vector is sampled per patch and copied over that patch.  The initialization routine performs one Flow-Lenia step before returning the initial state.  Exact replay therefore matches the returned post-step $A/P$ state rather than only the pre-step patch draws.

\paragraph{Rule decoding and update.}
The rule coordinates determine the global radius, relative kernel radii, growth centers and widths, radial centers, amplitudes and widths, and a food-consumption coordinate that is dynamically inactive because food is disabled.  Raw offsets are added to the base rule, passed through a sigmoid and mapped to the physical parameter ranges.  For input routing $c_0(\ell)$ and output-routing sets $\mathcal K_c$, the normalized radial kernels $K_\ell$ define
\[
  U_\ell(x)=(K_\ell*A_{t,c_0(\ell)})(x),
  \qquad
  G_\ell(x)=\left[2\exp\!\left(-\frac{(U_\ell(x)-m_\ell)^2}{2s_\ell^2}\right)-1\right]P_t(x,\ell),
\]
\[
  B_c(x)=\sum_{\ell\in\mathcal K_c}G_\ell(x).
\]
The transport field is formed from Sobel gradients,
\[
  \widetilde F_c=\nabla_S B_c,
  \qquad
  C_A=\nabla_S\sum_d A_{t,d},
\]
\[
  \alpha_c=\operatorname{clip}\!\left((A_{t,c}/2)^2,0,1\right),
  \qquad
  F_{t,c}=\operatorname{clip}\!\left((1-\alpha_c)\widetilde F_c-\alpha_c C_A,
  -(d_d-\sigma_{\rm RT}),d_d-\sigma_{\rm RT}\right).
\]
The proposed RT displacement additionally includes $dt$.  The RT operator redistributes each source cell over a box footprint centered at the proposed target and transports $P$ over the same footprints.  In categorical mixing mode, an incoming internal state is selected with probability proportional to its incoming mass.  The baseline simulator is intended as mass transport but does not explicitly renormalize mass after each step; wall handling and finite-grid reintegration can therefore produce long-horizon numerical mass loss.

The rendered frame used for videos and CLIP is
\[
  \operatorname{RGB}(x)=\operatorname{clip}\!\left(\left[\sum_c A(x,c)\right]P(x,0{:}3),0,1\right).
\]
Thus the visual readout depends on both transported mass and the first three internal fields.

\paragraph{Tracer observation layer.}
Each MSPD rollout tracks $8192$ passive tracers.  A tracer cell is sampled with probability proportional to total local activity, its within-cell position is uniformly jittered, and its channel is sampled from local channel proportions.  After each substrate step, the channel is resampled from updated local proportions, the channel flow is bilinearly sampled, and RT box diffusion adds
\[
  \varepsilon_a(t)\sim\operatorname{Unif}([ -\sigma_{\rm RT},\sigma_{\rm RT}]^2).
\]
The update is
\[
  q_a(t+1)=\Pi_\Omega\!\left(q_a(t)+\operatorname{clip}(dt\,v_a(t),-m_{\rm step},m_{\rm step})+\varepsilon_a(t)\right),
\]
where $\Pi_\Omega$ clips to the wall-bounded domain.  The logged coordinates \texttt{lagrangian\_xy} and optional channels \texttt{lagrangian\_c} define the nonperiodic finite-lag displacement laws used by MSPD.

\paragraph{Metric grid.}
Optimization and exact C1 replay use a $300{,}000$-step horizon, sampling every $50$ steps, metric range $50{,}000$--$300{,}000$, window extent $20{,}000$, stride $5{,}000$, and lags $1000$ through $10{,}000$ in $1000$-step increments.  This gives $47$ windows.  Each metric cell samples at most $48$ lag origins, $64$ tracers and $16$ fixed projection directions; six pooled-null replicates estimate finite-sample identity-free heterogeneity.  The processed trace is $\max(\Delta H,0)$ and the amplitude term is disabled.  The temporal MSPD score averages the four valid dyadic reconstruction errors for scales $1,2,4,8$ on the $47$-window trace.

\subsubsection*{B.10 Flow-Lenia pathwise stochasticity audit}
\label{app:flow-stochasticity-audit}

This auxiliary analysis motivates the fixed-context formulation used by the Flow-Lenia optimization.  It is not treated as an additional C1, C2 or C5 hypothesis.

\paragraph{Protocol.}
The experiment compares two sources of trajectory separation while holding the realised physical start state fixed.  It uses thirty random rule vectors drawn from the same initialization distribution as the matched controls, four bit-exact shared initial states and eight independent stochastic continuations per state and rule.  A duplicate lane exactly repeats the first continuation as an execution audit.  No external perturbation is added.  Within-rule divergence is computed over the $\binom{8}{2}=28$ continuation pairs for a fixed rule and initial state.  Between-rule divergence is computed over the $\binom{30}{2}=435$ rule pairs with both the initial state and continuation seed matched.  All trajectories use the same fixed batch topology and are continued for $300{,}000$ simulator steps.

The pointwise audit uses same-time CLIP cosine distance; the windowed audit uses symmetric CLIP--Chamfer over $20$ rendered frames in $20{,}000$-step windows.  Each within-rule curve is normalized by its own late between-rule reference.  Under both summaries, within-rule stochastic divergence reaches the between-rule scale rapidly and remains of comparable magnitude over the long rollout.

\begin{figure}[H]
\centering
\includegraphics[width=0.94\linewidth]{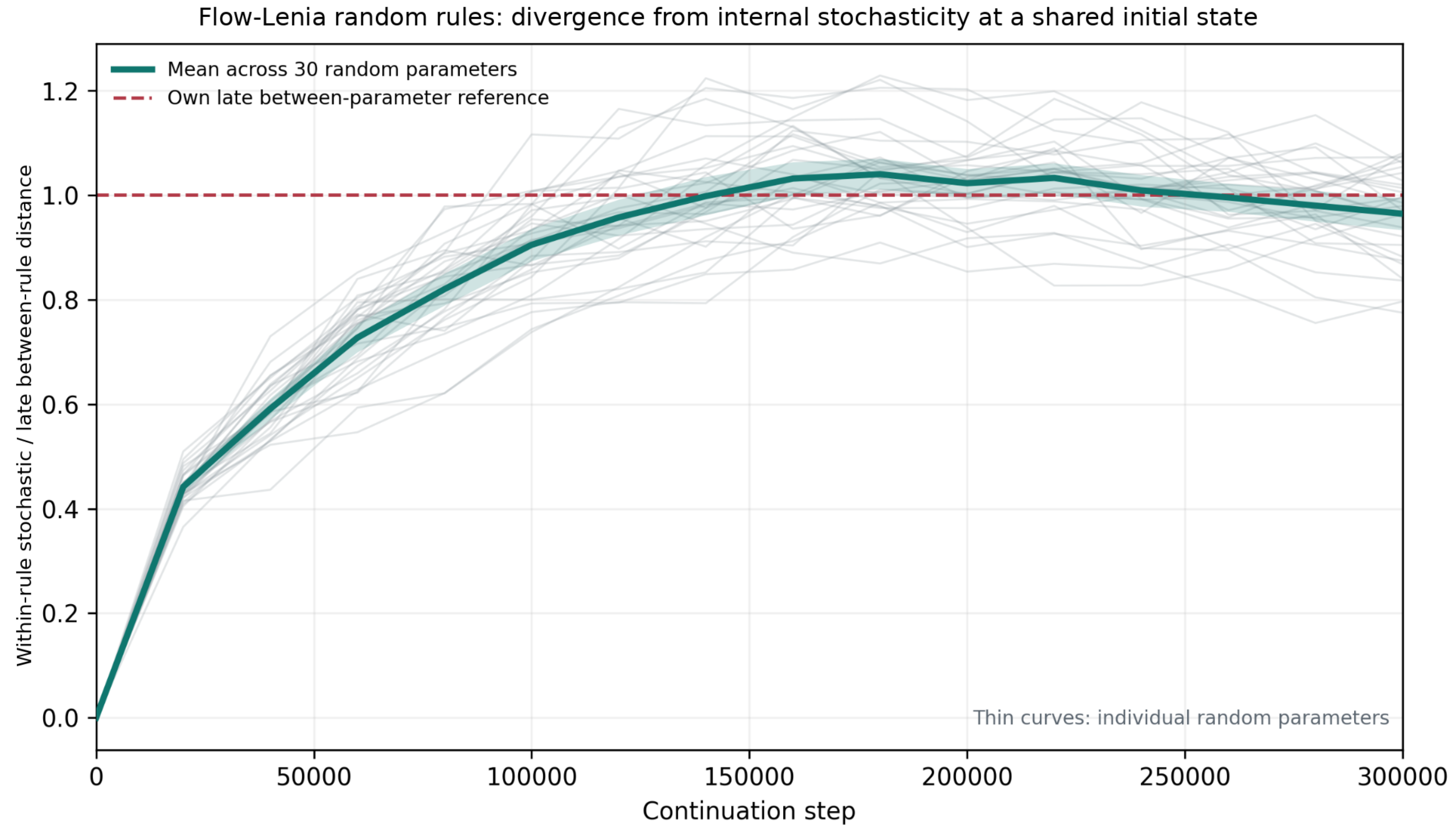}
\caption{Long-horizon exact-state stochasticity audit for random Flow-Lenia rules.  For each of thirty random parameter vectors, within-rule CLIP--Chamfer divergence between continuations with independent stochastic realizations is normalized by its late between-rule reference under matched initial states and continuation seeds.  Thin curves show individual random rules and the thick curve is their equal-weight mean.  The mean approaches the between-rule scale and remains near it through the full rollout.  This measures separation of finite stochastic trajectories under the fixed simulator protocol; it is not a deterministic Lyapunov analysis.}
\label{fig:flow-stochasticity-vs-checkpoint}
\end{figure}
\FloatBarrier

\paragraph{Interpretation.}
For random rules from the matched-control initialization distribution, divergence induced solely by the simulator's internal stochasticity can become comparable to divergence caused by changing the rule parameters.  The effect is therefore not restricted to optimized checkpoints.  Estimating an MSPD objective averaged over internal stochasticity would require substantially more independent continuations than the available optimization budget permits.  We consequently condition Flow-Lenia search and exact replay on realised pathwise contexts.  This audit measures finite stochastic trajectory separation and does not estimate a deterministic Lyapunov exponent.

\subsubsection*{B.11 Particle Life++ substrate algorithm}
\label{app:plife-plus-algorithm}

This appendix gives the algorithmic form of Particle Life++ used in Section~4.  The substrate is a deterministic particle system after initialization when the network weights, initial state and boundary rule are fixed.

\paragraph{State and initialization.}
At step $n$, the state is
\[
  S_n=\{(x_i^n,v_i^n,c_i^n)\}_{i=1}^{N},
  \qquad
  x_i^n\in[0,L]^2,
  \quad v_i^n\in\mathbb R^2,
  \quad c_i^n\in\mathbb S^{K-1}.
\]
The implementation initializes positions uniformly in the square, velocities at zero, and colors by normalizing independent Gaussian vectors,
\[
  x_i^0\sim\mathrm{Unif}([0,L]^2),
  \qquad
  v_i^0=0,
  \qquad
  c_i^0=\xi_i/\|\xi_i\|_2,
  \quad \xi_i\sim\mathcal N(0,I_K).
\]

\paragraph{Neural pair rule.}
For each ordered pair $(i,j)$, the network receives the concatenated colors $[c_i,c_j]\in\mathbb R^{2K}$.  The architecture is
\[
  \mathbb R^{2K}
  \xrightarrow{\mathrm{Dense}(8),\tanh}
  \mathbb R^8
  \xrightarrow{\mathrm{Dense}(8),\tanh}
  \mathbb R^8
  \xrightarrow{\mathrm{Dense}(8),\tanh}
  \mathbb R^8
  \xrightarrow{\mathrm{Dense}(1+K)}
  \mathbb R^{1+K}.
\]
If the final output is $(y_{ij}^{(0)},y_{ij}^{(1:K)})$, then
\[
  \alpha_{ij}=1.5\tanh(y_{ij}^{(0)}),
  \qquad
  g_{ij}=\tanh(y_{ij}^{(1:K)}).
\]
The scalar $\alpha_{ij}$ controls the mechanical interaction amplitude and $g_{ij}$ controls the color drift of particle $i$ under particle $j$.

\paragraph{Pair geometry and force law.}
Let $\delta_{ij}$ be the displacement from $i$ to $j$.  Under the toroidal boundary used in the reported PLife++ experiments, $\delta_{ij}$ is the minimal-image displacement; under wall boundaries it is the ordinary displacement.  Define
\[
  \rho_{ij}=\|\delta_{ij}\|_2,
  \qquad
  e_{ij}=\frac{\delta_{ij}}{\rho_{ij}+10^{-8}},
  \qquad
  q_{ij}=\rho_{ij}/r_{\max}.
\]
The dimensionless force graph is
\[
\Phi(q;\alpha,\beta)=
\begin{cases}
  q/\beta-1, & q<\beta,\\[1mm]
  \alpha\left(1-\dfrac{|2q-1-\beta|}{1-\beta}\right), & \beta<q<1,\\[2mm]
  0, & \text{otherwise}.
\end{cases}
\]
The pairwise force and color increment are
\[
  F_{ij}=r_{\max}\Phi(q_{ij};\alpha_{ij},\beta)e_{ij},
  \qquad
  \Delta c_{ij}=g_{ij}\max(0,1-\rho_{ij}/r_{\max}).
\]
The dense implementation evaluates all ordered pairs.  The self-pair gives zero mechanical force because $\delta_{ii}=0$, but it can contribute to color drift through $g_\theta(c_i,c_i)$.

\paragraph{Integration step.}
The acceleration and color drift are
\[
  a_i^n=\frac{1}{m}\sum_{j=1}^{N}F_{ij}^n,
  \qquad
  \dot c_i^n=\sum_{j=1}^{N}\Delta c_{ij}^n.
\]
With damping factor $\mu=0.5^{\Delta t/h}$, the semi-implicit update is
\[
  v_i^{n+1}=\mu v_i^n+\Delta t\,a_i^n,
  \qquad
  \widetilde x_i^{n+1}=x_i^n+\Delta t\,v_i^{n+1}.
\]
The boundary map then gives $x_i^{n+1}=\mathrm{boundary}(\widetilde x_i^{n+1})$ and updates $v_i^{n+1}$ only for wall reflections.  For toroidal boundaries, $x_i^{n+1}=\widetilde x_i^{n+1}\bmod L$ and velocity is unchanged.  If color updates are enabled,
\[
  \widetilde c_i^{n+1}=c_i^n+\Delta t\,\dot c_i^n,
  \qquad
  c_i^{n+1}=\widetilde c_i^{n+1}/\|\widetilde c_i^{n+1}\|_2.
\]
Particle identities $i=1,\ldots,N$ are the local carriers for MSPD, and the empirical local laws are built from finite-lag displacements of the stored positions $x_i^n$.

The PLife++ experiments use $N=600$, $K=6$, $d=2$, $\beta=0.3$, $m=0.1$, $\Delta t=0.02$, $h=0.04$, $r_{\max}=0.1$, $L=1.0$, toroidal boundary, dense neighbor mode, color updates enabled, render radius $0.04$, sharpness $30.0$ and black background.

\subsection{B.12 branch perturbation protocols}
\label{app:c2-branch-implementation}

C2 uses substrate-native controlled continuation families.  The rule parameters are never changed.  CA modifies the deterministic board state; Flow-Lenia preserves the exact state and samples independent stochastic continuations.  The values below are the explicit parameters used for the reported experiments.

\paragraph{Cellular automata.}
A CA branch state is a binary board sampled from the selected rule trajectory.  For each selected branch window, the branch time is the center of the window.  Each continuation is initialized by independently choosing a random subset of board sites and flipping their binary states:
\[
  x'(u)=
  \begin{cases}
  1-x(u), & u\in S_{\mathrm{flip}},\\
  x(u), & u\notin S_{\mathrm{flip}},
  \end{cases}
  \qquad
  |S_{\mathrm{flip}}|=\max\{1,\operatorname{round}(0.01|\Lambda|)\}.
\]
The same Life-like rule is then resumed from $x'$.  The reported CA C2 parameters are
\[
  n_{\mathrm{windows}}=24,\qquad
  M=4,\qquad
  \text{flip fraction}=0.01,\qquad
  \text{future horizon}=64\ \text{steps}.
\]
The outcome distance is normalized Hamming divergence between future board states, averaged over the resumed horizon.

\paragraph{Flow-Lenia.}
A Flow-Lenia branch state contains the exact saved activity field, internal field, tracer state, simulator stochastic state and optimizer-native execution context.  For each selected state, three continuations are created with independent continuation seeds while preserving the saved state.  The external perturbation scales are zero, so $A$, $P$ and tracer coordinates match the source state exactly at branch offset zero.

The ten source trajectories contribute five high-, five middle- and five low-energy branch states each.  High and low are the top and bottom $20\%$ of within-trajectory energy ranks; middle is the $40\%$--$60\%$ band.  The minimum branch step is $50{,}000$, selected times are separated by at least $5{,}000$ steps, and all requested times are exact saved snapshots.  Each branch is simulated to $30{,}000$ steps and scored on the five prefix horizons described in Appendix~B.6.  The common branch frame is excluded from every future cloud.

Figure~\ref{fig:c2-branch-selection-preview} shows the ten source trajectories and the selected high-, middle- and low-energy branch times used by the revised C2 analysis.

\subsubsection*{Appendix C. Optimization and post-hoc evaluation protocols}
\label{app:optimization-protocols}

This appendix records the optimization protocols used for the three substrates in Section~4.  In all cases, the optimizer or selector produces a parameter object $\theta$, while the reported C1 numbers are computed by a post-hoc MSPD protocol applied equally to optimized and matched random controls.

\paragraph{Life-like cellular automata.}
The optimized object is the 18-bit totalistic Life-like rule code.  The search evaluates all $2^{18}$ birth/survival rules on a periodic $64\times64$ binary board.  The sweep uses rollout length $T=2048$, burn-in $512$, window size $64$, window step $16$, active-cell sample cap $256$, Jensen--Shannon transition-law distance, pair sample $512$, one null repetition, initial density sampled from $[0.05,0.4]$, eight random initial boards per rule, random seed $0$, JAX metric batch size $8$ and evaluation batch size $512$.

For C1, the optimized candidate is the top rule from the completed sweep.  It is re-evaluated against $32$ random Life-like rules on $32$ fresh Bernoulli holdout boards drawn from the same initialization-density law.  The CA transition symbol is the categorical before-after neighborhood event
\[
  a_t(i)=\operatorname{code}(x_t|_{i+B},x_{t+1}(i)),
\]
with active-support weighting as described in Section~4.1.

\paragraph{Flow-Lenia.}
The optimizer-updated rule component is a vector of $110$ raw-logit offsets $\phi$ around a fixed base rule.  Optimization also includes one nuisance coordinate $u_\tau$ for the MSPD lag, giving the search vector
\[
  \vartheta=(\phi,u_\tau)\in\mathbb R^{111}.
\]
The lag coordinate is mapped by a sigmoid and rounding operation to the grid $\{1000,2000,\ldots,10000\}$ simulator steps.  It is part of the search vector but not part of the decoded dynamical rule.

There are ten independent optimization runs.  Run $r$ uses optimization seed $1003+r$ and four fixed evaluation keys with bases $400003+2r$.  The four keys are reused at every iteration.  Candidate evaluation uses the canonical $128\times128$ wall-bounded simulator, a $300{,}000$-step rollout, sampling every $50$ steps, metric range $50{,}000$--$300{,}000$, window size $20{,}000$, window step $5{,}000$, $8192$ logged tracers, $48$ sampled lag origins, $64$ sampled tracers, $16$ projection directions and six pooled-null replicates.  The objective is the mean MSPD over the four fixed realised contexts at the candidate's selected lag; the amplitude term is zero.

The campaign uses mirrored OpenAI-ES.  At each of $100$ iterations, four standard-normal directions generate eight antithetic candidates $\vartheta_i\pm0.2\epsilon_j$.  Each candidate is evaluated on the same four fixed contexts.  Scores are standardized separately within each context column, the antithetic score differences form an ascent estimate, and Adam updates the center with learning rate $0.2$, $\beta_1=0.9$, $\beta_2=0.999$ and $\epsilon=10^{-8}$.  The initial center is zero and receives no additional initialization noise.

The paper candidate is selected after the complete sweep by simple global argmax over all $100\times8$ evaluated population members, using observed mean MSPD over the four fixed contexts.  There is no lower-confidence-bound or multi-stage stability selection.  The selected candidate is indexed directly from the stored population trajectory rather than from the smoothed center record.

\begin{table}[H]
\centering
\footnotesize
\begin{tabular}{rrrr}
\toprule
Run & Iteration & Population index & Selected lag \\
\midrule
000 & 73 & 6 & 10000 \\
001 & 87 & 4 & 10000 \\
002 & 31 & 3 & 8000 \\
003 & 56 & 0 & 7000 \\
004 & 38 & 4 & 8000 \\
005 & 49 & 6 & 10000 \\
006 & 79 & 3 & 9000 \\
007 & 45 & 3 & 7000 \\
008 & 98 & 3 & 10000 \\
009 & 91 & 3 & 10000 \\
\bottomrule
\end{tabular}
\caption{Flow-Lenia paper candidates selected by global observed-MSPD argmax over each completed OpenAI-ES sweep.}
\label{tab:flow-selected-candidates}
\end{table}
\FloatBarrier

C1 replays the selected rule on the same four exact contexts using the original population-by-context JIT/batch topology and key schedule.  The full replay contains $47$ metric windows.  Even-indexed split A contains $24$ windows and is retained as a diagnostic; odd-indexed split B contains $23$ windows and supplies the reported score.  The group lag is inherited from the optimized search vector and applied to the optimized rule and all three matched random rules.  Split B is therefore a temporally disjoint reporting partition, not fresh-seed or fresh-initial-state generalization.

\paragraph{Particle Life++.}
Particle Life++ optimization uses Sep-CMA-ES on the neural pair-interaction weights $\theta$ of $f_\theta(c_i,c_j)$.  The optimization substrate uses $600$ particles, $6$ colors, two spatial dimensions, $\beta=0.3$, mass $0.1$, $\Delta t=0.02$, half-life $0.04$, interaction radius $r_{\max}=0.1$, render radius $0.04$, sharpness $30.0$, color updates enabled, world size $1.0$, toroidal boundary, dense neighbor mode and black background.  The rollout length during optimization is $6000$ steps with sampling every step.

The optimizer uses substrate-default parameter initialization, batch size $1$, population batch $8$, population size $8$, $500$ iterations and initial standard deviation $0.2$.  The optimization-time metric reads particle positions \texttt{state\_x}, unwraps toroidal positions, uses trainable lag grid $\{5,10,15,20,25,30,35\}$, window size $100$, window step $50$, metric range $0$--$1000$, $48$ displacement samples, minimum $4$ samples, $16$ projections, six null repetitions, $64$ sampled particles, preprocessing mode \texttt{clip}, $\alpha=1$, $\beta=1$ and $\epsilon=10^{-3}$.  The objective includes both the MSPD term and a mean-heterogeneity term, which prevents CMA-ES from remaining in low-motion stationary basins of random neural weights.

The reported PLife++ C1 score is computed post hoc from reusable single-trajectory rollouts rather than directly from optimization-time scores.  The post-hoc dynamics use the same substrate profile as the C5 simulation, total horizon $24{,}000$ steps and sampling every step.  MSPD is evaluated on steps $4000$--$24{,}000$ with fixed $\tau=25$, window size $100$, window step $50$, toroidal unwrapping, $48$ samples, $16$ projections, six null repetitions and $64$ sampled particles.  Seven optimized groups are compared with three matched random baselines per group.

\paragraph{Matched random controls.}
Flow-Lenia has three matched random rule controls per optimization run.  They are sampled independently from the same zero-centered initialization law and scale using the first ask population of an eight-member Sep-CMA-ES sampler with group seed $200003+10000r$; members $0,1,2$ are retained.  They are not OpenAI-ES iteration-zero candidates.  The random rules have no optimized lag coordinate and are evaluated at the lag selected by the optimized candidate in their group.  During C1 replay, each random rule retains its own decoded rule-dependent fields but starts from the exact optimized candidate's returned physical state for the corresponding context, followed by matched tracer initialization.

Particle Life++ matched random checkpoints retain the existing construction from the substrate-default Sep-CMA-ES initialization distribution.  For matched group $r$ and random index $j$, the random seed is
\[
  s_{r,j}=200001+10000r+\lfloor j/P\rfloor,
\]
where $P=8$ and $j\bmod P$ selects the member from the first generated population.  These controls are simulated and scored by the same Particle Life++ post-hoc protocol as the optimized checkpoints.

\paragraph{Image encoder used for rollout-distance features.}
\label{app:encoder-description}
The representation-space distances used in C2 and C5 use the same fixed image-encoder backend as the ASAL-style CLIP evaluation.  In all reported CLIP--Chamfer computations, $f$ is the ASAL \texttt{clip} backend: OpenAI CLIP ViT-B/32, loaded as \texttt{openai/clip-vit-base-patch32} through the HuggingFace \texttt{FlaxCLIPModel} and \texttt{AutoProcessor}.  The encoder is frozen and is not trained on the generated rollouts.

Each rendered RGB frame $R(t)$ is represented as an array with values in $[0,1]$.  If necessary, it is resized to $224\times224$ by bilinear interpolation, normalized by the CLIP image-processor mean and standard deviation, and passed through the CLIP image tower.  The resulting image feature is normalized to unit Euclidean norm:
\[
  f(R)
  =
  \frac{\operatorname{CLIP}_{\mathrm{img}}\!\left(
  \operatorname{norm}_{\mu,\sigma}\!\left(
  \operatorname{resize}_{224}(R)
  \right)\right)}
  {\left\|
  \operatorname{CLIP}_{\mathrm{img}}\!\left(
  \operatorname{norm}_{\mu,\sigma}\!\left(
  \operatorname{resize}_{224}(R)
  \right)\right)
  \right\|_2}.
\]
Thus all frame embeddings used in the Chamfer computation lie on the unit sphere, and pairwise frame distance is the cosine distance $d_{\cos}(u,v)=1-u^\top v$.  This encoder is used only to measure future-rollout outcome divergence in C2 and C5; it is not part of the MSPD definition or the MSPD optimization objective.

\subsubsection*{Appendix D. Rendered synthetic calibration examples}
\label{app:synthetic-rendered-examples}

\begin{figure}[p]
  \centering
  \includegraphics[width=\textwidth,height=0.82\textheight,keepaspectratio]{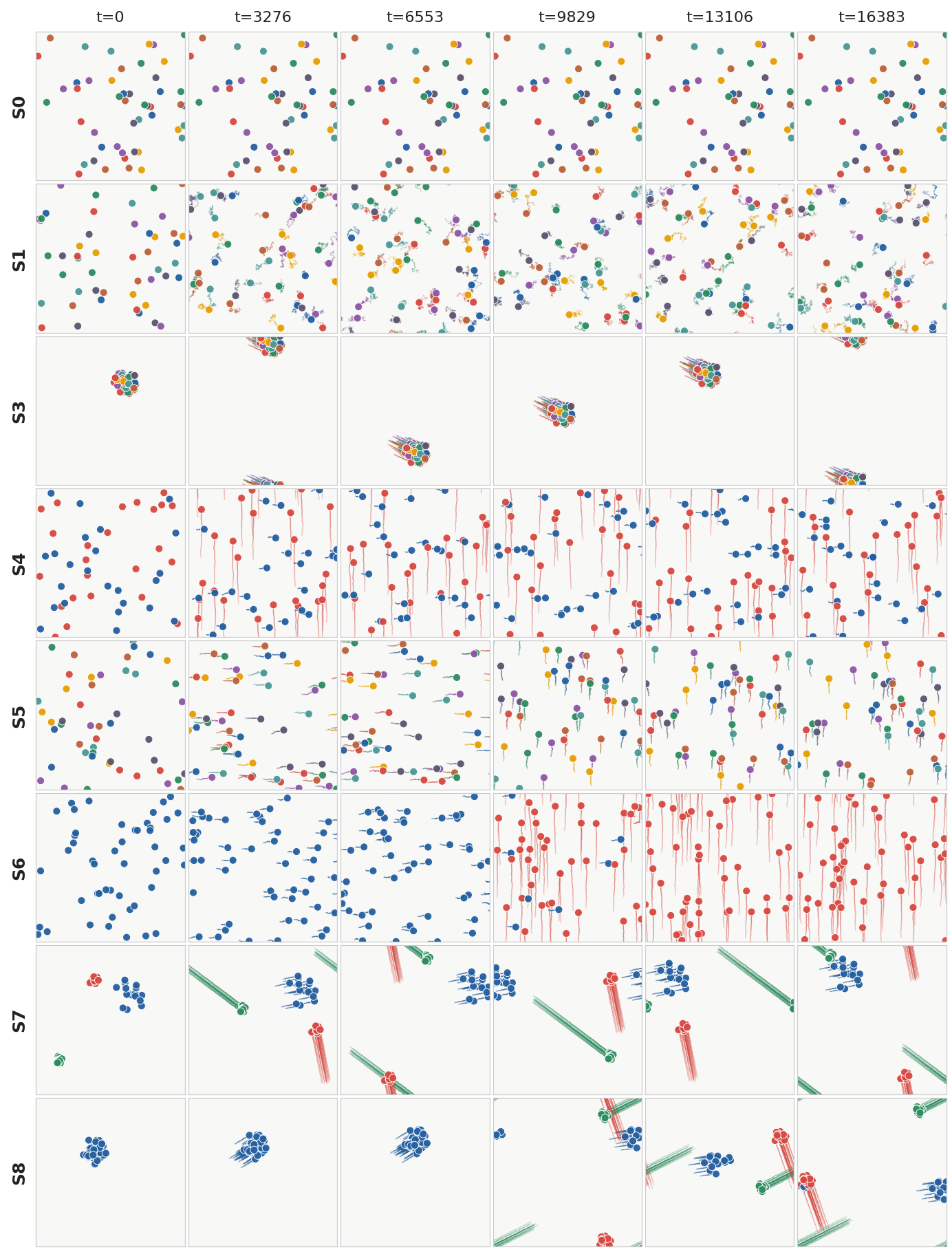}
  \caption{
  Rendered examples from the synthetic calibration suite.  Rows correspond to synthetic families S0--S8 and columns show selected rollout times.  These renderings illustrate the qualitative regimes used in N0; the metric itself is evaluated on trajectory-derived transition laws.
  }
  \label{fig:synthetic-frame-montage-appendix}
\end{figure}

Figure~\ref{fig:synthetic-frame-montage-appendix} shows rendered examples from the synthetic calibration suite.  The figure is included only as a visual reference for the controlled cases; the N0 MSPD scores are computed from particle trajectories and finite-lag displacement laws, not from rendered frames.

\end{document}